\documentclass[final,3p,times]{elsarticle_tech}
\usepackage{amsmath}  
\usepackage{overpic,rotating,contour,subfigure}                         
\usepackage{graphics,subfigure}
\usepackage{graphicx}
\usepackage{epsfig}
\usepackage{amssymb}  
\usepackage{epsfig} 
\usepackage{psfrag,rotating}
\usepackage{amssymb,floatflt,enumerate}
\usepackage{amsmath,amscd,psfrag,leqno} 
\usepackage[mathscr]{eucal}   
\usepackage{shadethm}   
\usepackage{graphicx,subfigure}
\usepackage{overpic,contour}  
    
\def\Beweisende{\square}            
\def\BewEnde{\hfill{\Beweisende}}

\def\phm{{\hphantom{-}}} 

\def\phi{\varphi}

\def\RR{{\mathbb R}}

\def\CC{{\mathbb C}}

\def\Vkt#1{{\mathbf #1}} 

\newcommand{\go}[1]{{\sf #1}}

\newtheorem{thm}{Theorem}

\newtheorem{definition}{Definition}

\newdefinition{rmk}{Remark}
\newdefinition{cor}{Corollary}

\begin{document}
 
\begin{frontmatter}
 
\title{Pentapods with Mobility 2}

\author[TU]{Georg Nawratil}
\author[RADON]{Josef Schicho}
\address[TU]{Institute of Discrete Mathematics and Geometry, Vienna University of Technology, 
Wiedner Hauptstrasse 8-10/104, 1040 Vienna, Austria}
\address[RADON]{Johann Radon Institute for Computational and Applied Mathematics, Austrian Academy of Sciences, 
Altenberger Strasse 69,
4040 Linz, Austria}
    
\begin{abstract} 
In this paper we give a full classification of all pentapods with mobility 2, where neither all platform anchor points nor 
all base anchor points are located on a line. 
Therefore this paper solves the famous Borel-Bricard problem for 
2-dimensional motions beside the excluded case of five collinear points with spherical trajectories. 
But even for this special case we present three new types as a side-result.
Based on our study of pentapods, we also give a complete list of all non-architecturally singular hexapods
with 2-dimensional self-motions.
\end{abstract}

  
\end{frontmatter}

\section{Introduction}\label{intro}

\begin{figure}[h!]
\begin{center}  
 \begin{overpic}
    [width=55mm]{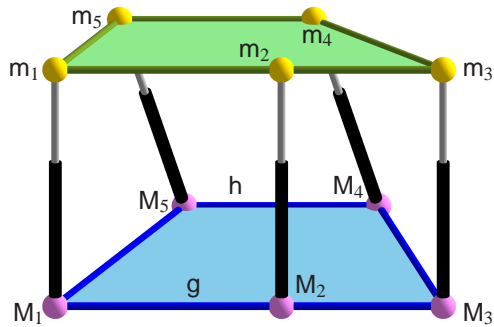}
\put(-6.5,0){$\go M_1$}
\put(61,6.5){$\go M_2$}
\put(101,0){$\go M_3$}
\put(70,30.5){$\go M_4$}
\put(23.5,27){$\go M_5$}
\put(-7.5,59){$\go m_1$}
\put(47.5,63.5){$\go m_2$}
\put(35,7){$\go g$}
\put(101,59){$\go m_3$}
\put(62.5,67){$\go m_4$}
\put(7.5,71.5){$\go m_5$}
\put(45,30.5){$\go h$}
  \end{overpic} 
\end{center}
\caption{Sketch of a pentapod with planar platform and planar base, which is referred as planar pentapod. 
Moreover this planar pentapod has a 2-dimensional self-motion due to its special geometric design (cf.\ item 2 of Theorem \ref{thm3}).}
  \label{fig0}
\end{figure}

The geometry of a pentapod (see Fig.\ \ref{fig0}) is given by the five base anchor points $\go M_i$ with coordinates
$\Vkt M_i:=(A_i,B_i,C_i)^T$ with respect to the fixed system and by the five platform anchor points $\go m_i$ 
with coordinates $\Vkt m_i:=(a_i,b_i,c_i)^T$ with respect to the moving system (for $i=1,\ldots ,5$). 
Each pair $(\go M_i,\go m_i)$ of corresponding anchor points is connected by a SPS-leg, where only 
the prismatic joint (P) is active and the spherical joints (S) are passive. 

If the geometry of the manipulator is given, as well as the lengths of the five pairwise distinct legs, the 
pentapod has generically mobility 1 according to the formula of Gr\"ubler. The corresponding motion is called a 
1-dimensional self-motion of the pentapod. 
But, under particular conditions, the manipulator can 
gain additional mobility.  
These pentapods represent interesting solutions to the 
still unsolved problem posed by the French Academy of Science for the {\it Prix Vaillant} of the year 1904, which is also known as 
Borel-Bricard problem (cf.\ \cite{borel,bricard,husty_bb}) and reads as follows: 
{\it "Determine and study all displacements of a rigid body in which distinct points of the body move on spherical paths."}


\subsection{Related results}

The classification of pentapods with mobility 2 is based on the following theorem 
proven in \cite{gns2}. 

\newpage

\begin{thm}\label{thm1a}
If the mobility of a pentapod is 2 or higher, then one of the following conditions holds
\footnote{After a possible necessary renumbering of anchor points and exchange of the platform and the base.}:
        \begin{enumerate}[(a)]
        \item  
        The platform and the base are similar. This is a so-called equiform pentapod.
        \item
        The platform and the base are planar and affine equivalent. This is a so-called planar affine pentapod.
        \item
        There exits $p\leq 5$ such that $\go m_1,\ldots,\go m_p$ are collinear and 
        $\go M_{p+1},\ldots ,\go M_5$ are equal; i.e.\  $\go M_{p+1}=\ldots =\go M_5$.
        \item
        $\go M_1,\go M_2,\go M_3$ are located on the line $\go g$ which is parallel to the line $\go h$ spanned 
				by $\go M_4$ and $\go M_5$. Moreover $\go m_1,\go m_2,\go m_3$ are located on the line $\go g^{\prime}$ which is parallel to the line $\go h^{\prime}$ spanned 
				by $\go m_4$ and $\go m_5$.
        \end{enumerate}
\end{thm} 

We can focus on pentapods with 2-dimensional self-motions, as those with higher-dimensional ones follow from 
Theorem 7 of \cite{nawratilmulti}.  
They all belong to item (c) of Theorem \ref{thm1a} and read as follows (for more details please see \cite{nawratilmulti}):

\begin{cor}\label{thm1b}
A pentapod has an $n$-dimensional self-motion with $n>2$ in one of the following three 
cases (under consideration of footnote 1):
\begin{enumerate}[(i)]
\item
$\go m_1=\go m_2=\go m_3$ and $\go M_4=\go M_5$.
\item
$\go m_1=\go m_2=\go m_3=\go m_4$.
\item
All base points are collinear, all platform points are collinear and 
corresponding anchor points are related by a regular projectivity: $\go m_i\mapsto \go M_i$ for $i=1,\ldots ,5$. 
\end{enumerate}
\end{cor}
 
Moreover, we can restrict to the case $p\leq 4$ 
in item (c) of Theorem \ref{thm1a}, as we only study pentapods where neither all platform anchor points nor 
all base anchor points are collinear. The reason for excluding $p=5$ is that it 
requires a special treatment, as a 1-dimensional self-motion of the carrier line $\go g$ of 
$\go m_1,\ldots ,\go m_5$ already implies mobility 2 of the moving space, as in each pose of $\go g$ 
a 1-dimensional rotation about it is possible. Therefore the self-motions of pentapods with linear platform are 
studied separately in \cite{ns}.

In the following we list all pentapods\footnote{In the remainder of the article the word pentapod denotes 
a 5-legged manipulator  with non-collinear platform points and non-collinear base points; exceptions are noted explicitly.}
with mobility 2, which are known in the literature until now,
with respect to the three items of Theorem \ref{thm1a}:

\paragraph*{Examples ad (a), (b) and (d)}
The platform and the base are congruent (= so-called congruent pentapod) and all legs have the same length.
This pentapod, which belongs to case (a), can perform a 2-dimensional translational self-motion.

If the platform and base have the additional property to be planar, then we get an example
which also belongs to item (b). In the special case, where the anchor points are located on two parallel lines,
the pentapod possesses an additional 2-dimensional self-motion,
which is neither pure translational nor pure spherical (cf.\ Example 2 of \cite{nawratilmulti} 
after the removal of one leg). This example also belongs to item (d) of Theorem \ref{thm1a}.

\paragraph*{Examples ad (c)}
The remaining known examples belong to this class and read as follows (under consideration of footnote 1):

\begin{enumerate}[$\bullet$]
  			\item
  			Architecturally singular\footnote{A pentapod (hexapod) is called architecturally singular if in any pose of the platform
  			the rank of its Jacobian matrix $\Vkt J$ is less than five (six), which is equivalent with the statement that the carrier lines of the five (six) legs
  			belong to a linear congruence of lines (linear line complex). This equivalence is easy to see, as $\Vkt J$ is composed of the Pl\"ucker coordinates of 
  			these five (six) lines (cf.\ \cite{pottmann_wallner}).}
  			pentapods have at least a 2-dimensional self-motion in each pose over $\mathbb C$ as they are redundant.
  			For the listing of these manipulators see Theorem 3 of \cite{kargernonplanar} under consideration of  \cite{nawratilpenta}. 
  			As all these cases are known, they are not of further interest. 
  			In this context it should also be noted that the designs (ii) and (iii) of Corollary \ref{thm1b} are also architecturally singular ones. 
   			\item
        $\go m_1=\go m_2$ and $\go M_3=\go M_4$: If the platform is placed in a way that
        $\go m_1=\go m_2$ coincides with $\go M_3=\go M_4$, then there exists a 2-dimensional spherical self-motion (cf.\ \cite{nawratilmulti}).
        \item
        For $p=2$ the condition given in item (c) of Theorem \ref{thm1a} is already sufficient, as there exists a 2-dimensional spherical self-motion
        if the platform is placed in a way that $\go m_1$ or $\go m_2$ coincide with $\go M_3=\go M_4=\go M_5$ (cf.\ \cite{nawratilmulti}). 
        Note that we get architecturally singular pentapods for $p<2$ (cf.\ item (ii) of Corollary \ref{thm1b}). 
\end{enumerate}

\subsection{Results and outline of the article}

In this paper we study the three cases given in Theorem \ref{thm1a} in more detail. 
The obtained results are summarized in the following three theorems: 

\begin{thm}\label{thm2}
A non-planar equiform pentapod can only have a 2-dimensional self-motion
in the special case of congruent platform and base: The 2-dimensional self-motion 
is the trivial 2-dimensional translation. 
\end{thm}

\begin{thm}\label{thm3}
A planar affine pentapod has mobility 2 
in one of the following two cases (under consideration of footnote 1), if the pentapod is assumed to be not architecturally singular ($\Leftrightarrow$ no 4 points are collinear): 
\begin{enumerate}
\item
The affinity is a congruence: In this case the 2-dimensional self-motion is pure translational (planar case of Theorem \ref{thm2}).
\item
$\go M_1,\go M_2,\go M_3$ are located on the line $\go g$ which is parallel to the line $\go h$ spanned 
by $\go M_4$ and $\go M_5$. Moreover the restriction of the affinity to the lines is a congruence transformation. 
The 2-dimensional self-motion is a Sch\"onflies motion, where the axis of rotation is parallel to $\go g,\go h$ and 
the corresponding two lines in the platform (see Fig.\ \ref{fig0}).
\end{enumerate}
\end{thm}

\begin{thm}\label{thm4}
A non-architecturally singular pentapod with mobility 2, which is not listed in 
Theorems \ref{thm2} and \ref{thm3}, has to be one of the following designs (under consideration of footnote 1):
\begin{enumerate}
\item
$p=2$: The given condition $\go M_3=\go M_4=\go M_5$  is already sufficient for mobility 2. The 2-dimensional spherical self-motion 
is obtained if $\go M_3=\go M_4=\go M_5$ coincides with $\go m_1$ or $\go m_2$. 
\item
$p=3$; i.e.\ $\go M_4=\go M_5$ and $\go m_1,\go m_2,\go m_3$ are collinear: In the following two cases we get mobility 2:
	\begin{enumerate}
	\item
	$\go m_2=\go m_3$:  If $\go M_4=\go M_5$ coincides with $\go m_2=\go m_3$ we obtain the 2-dimensional spherical self-motion.
	\item
	$\go M_2,\go M_3,\go M_4=\go M_5$ collinear: If $\go M_4=\go M_5$ coincides with $\go m_1$ we obtain the 2-dimensional spherical self-motion.  
	\end{enumerate}
\item
$p=4$; i.e.\ $\go m_1,\ldots ,\go m_4$ are collinear:  
For $\go M_2,\go M_3,\go M_4,\go M_5$ collinear we get mobility 2. The 2-dimensional spherical self-motion is obtained if 
$\go m_1$ coincide with $\go M_5$. 
\end{enumerate}
\end{thm}

This complete classification of pentapods with mobility 2 reveals new cases, which can easily be seen by comparing the listed designs with 
the known results. It should be mentioned that item 2 of  Theorem \ref{thm3} is the generalization of the already mentioned Example 2 of \cite{nawratilmulti}. 
Note that this example is the only one, which possesses both 2-dimensional self-motions given in Theorem \ref{thm3}.

\begin{rmk}\label{rmk1}
As item 2 of  Theorem \ref{thm3} is known now, its existence can also easily be argued from 
the following property implied by the Sch\"onflies motion group (cf.\ \cite{nawratil_proj}): Every leg 
can be translated in direction of the rotation axis of the Sch\"onflies self-motion 
without changing this motion. 
Therefore the five legs only imply two constraints to the 4-dimensional  Sch\"onflies motion group. 
From this point of view this solution is also trivial.
\hfill $\diamond$
\end{rmk}

The proof of Theorem \ref{thm2} and \ref{thm3}, which is 
given in Section \ref{sec:pthm2} and \ref{sec:pthm3}, respectively, is based on the 
theory of bonds presented in Section \ref{sec:bon}. 
The proof of Theorem \ref{thm4} is given in Section \ref{sec:pthm4} and consists of two parts: 
In the first part (cf.\ Sections \ref{part(c)}), item (c) of Theorem \ref{thm1a} is studied by pure geometric-kinematic considerations/arguments, where 
also new results on self-motions of pentapods with linear platform are obtained as a side-result.  
In the second part (cf.\ Sections \ref{part(d)} and Appendix), item (d) of Theorem \ref{thm1a} is again discussed by means of bond theory. 

Based on Theorems \ref{thm2}, \ref{thm3} and \ref{thm4} we give a complete list of all non-architecturally singular hexapods (cf.\ footnote 3) with 
mobility 2 in Section \ref{sec:hexapod}.

\section{Bond Theory} \label{sec:bon}

In Section \ref{sec:def}, we give a short introduction to the theory of bonds for pentapods presented in \cite{nawratil_bond}, 
which was motivated by the bond theory of overconstrained closed linkages with revolute joints given in  \cite{hegedus} (see also \cite{hegedus2}). 
We start with the direct kinematic problem of pentapods and proceed with the definition of bonds. 
Based on these basics, we do some preparatory work in Section \ref{sec:class} by giving a classification of  2-dimensional self-motions, 
which is induced by the bond theory in a natural way.

\subsection{Definition of bonds}\label{sec:def}

Due to the result of Husty \cite{husty}, it is advantageous to work with 
Study parameters $(e_0:e_1:e_2:e_3:f_0:f_1:f_2:f_3)$ for solving the forward kinematics. 
Note that the first four homogeneous coordinates $(e_0:e_1:e_2:e_3)$ are the so-called Euler parameters.
Now, all real points of the Study parameter space $P^7$ (7-dimensional projective space), 
which are located on the so-called Study quadric $\Psi:\,\sum_{i=0}^3e_if_i=0$, 
correspond to an Euclidean displacement, with exception of the 3-dimensional subspace $E$ of $\Psi$ given by $e_0=e_1=e_2=e_3=0$, 
as its points cannot fulfill the condition $N\neq 0$ with $N=e_0^2+e_1^2+e_2^2+e_3^2$. 
The translation vector $\Vkt t:=(t_1,t_2,t_3)^T$ and the rotation matrix $\Vkt R$ of the corresponding 
Euclidean displacement $\Vkt x\mapsto\Vkt R\Vkt x + \Vkt t$ are given by:  
\begin{equation*}
t_1=2(e_0f_1-e_1f_0+e_2f_3-e_3f_2), \quad
t_2=2(e_0f_2-e_2f_0+e_3f_1-e_1f_3), \quad
t_3=2(e_0f_3-e_3f_0+e_1f_2-e_2f_1),
\end{equation*}  
and
\begin{equation}\label{mat:R}
\Vkt R = \begin{pmatrix} 
e_0^2+e_1^2-e_2^2-e_3^2 & 2(e_1e_2-e_0e_3) & 2(e_1e_3+e_0e_2)  \\
2(e_1e_2+e_0e_3) & e_0^2-e_1^2+e_2^2-e_3^2 & 2(e_2e_3-e_0e_1)  \\
2(e_1e_3-e_0e_2) & 2(e_2e_3+e_0e_1) & e_0^2-e_1^2-e_2^2+e_3^2
\end{pmatrix},  
\end{equation}  
if the normalizing condition $N=1$ is fulfilled. All points of the complex extension of $P^7$, which cannot fulfill this normalizing condition,  
are located on the so-called exceptional cone $N=0$ with vertex $E$. 

By using the Study parametrization of Euclidean displacements, the condition that the point $\go m_i$ is located 
on a sphere centered in $\go M_i$ with radius $R_i$ is a quadratic homogeneous equation according to Husty \cite{husty}.  
This so-called sphere condition $\Lambda_i$ has the following form: 
\begin{equation}\label{eq:lambda}
\begin{split}
\Lambda_i:\quad 
&(a_i^2+b_i^2+c_i^2+A_i^2+B_i^2+C_i^2-R_i^2)N 
-2(a_iA_i+b_iB_i+c_iC_i)e_0^2
-2(a_iA_i-b_iB_i-c_iC_i)e_1^2 \\
&+2(a_iA_i-b_iB_i+c_iC_i)e_2^2
+2(a_iA_i+b_iB_i-c_iC_i)e_3^2 
+4(c_iB_i-b_iC_i)e_0e_1
-4(c_iA_i-a_iC_i)e_0e_2 \\
&+4(b_iA_i-a_iB_i)e_0e_3 
-4(b_iA_i+a_iB_i)e_1e_2
-4(c_iA_i+a_iC_i)e_1e_3
-4(c_iB_i+b_iC_i)e_2e_3 \\
&+4(a_i-A_i)(e_0f_1-e_1f_0)
+4(b_i-B_i)(e_0f_2-e_2f_0) 
+4(c_i-C_i)(e_0f_3-e_3f_0) 
+4(a_i+A_i)(e_3f_2-e_2f_3) \\
&+4(b_i+B_i)(e_1f_3-e_3f_1)
+4(c_i+C_i)(e_2f_1-e_1f_2) 
+4(f_0^2+f_1^2+f_2^2+f_3^2) =0.
\end{split}
\end{equation}

Now the solution for the direct kinematics over $\CC$ of a pentapod can be written as the
algebraic variety $V$ of the ideal spanned by $\Psi,\Lambda_1,\ldots ,\Lambda_5,N=1$.
In general $V$ consists of a 1-dimensional set of points.

We consider the algebraic motion of the pentapod, which is defined as the set of points on the
Study quadric determined by the constraints; i.e.\ the  common points of  
the six quadrics $\Psi,\Lambda_1,\ldots ,\Lambda_5$. 
Now the points of the algebraic motion with $N\neq 0$ equal the kinematic image of the algebraic variety $V$. 
But we can also consider the points of the algebraic motion, which belong to the exceptional cone $N=0$. 
An exact mathematical definition of these so-called bonds can be given as follows (cf.\ Definition 1 of \cite{nawratil_bond}):  

\begin{definition}\label{df:bond5}
For a pentapod the set $\mathcal{B}$ of bonds is defined as:
\begin{equation*}
\mathcal{B}:=
ZarClo\left(V^{\star}\right)\,\,\cap\,\,
\left\{
(e_0:\ldots :f_3)\in P^7\,\, | \,\,  \Psi,\Lambda_1,\ldots ,\Lambda_5 , N=0 
\right\},
\end{equation*}
where $V^{\star}$ denotes the variety $V$ after the removal of all components, which correspond to pure translational motions.
Moreover $ZarClo(V^{\star})$ is the Zariski closure of $V^{\star}$, i.e.\  the zero locus of all 
algebraic equations that also vanish on $V^{\star}$. 
\end{definition}

The restriction to non-translational motions is caused by the following approach used for the computation of bonds: 
In a first step we project the algebraic motion of the pentapod 
into the Euler parameter space $P^3$ by the elimination of $f_0,\ldots ,f_3$. 
This projection is denoted by $\pi_f$. 
In a second step we determine 
those points of the projected point set $\pi_f(V)$, which are located on the quadric $N=0$; i.e.\ 
\begin{equation}\label{def:projbond}
\mathcal{B}_f:= ZarClo\left(\pi_f\left(V\right)\right) \cap \left\{
(e_0:\ldots :e_3)\in P^3\,\, | \,\,  N=0 
\right\}.
\end{equation}
Note that this set of projected bonds, which is denoted by $\mathcal{B}_f$, cannot be empty for an non-translational self-motion.

Clearly, the kernel of this projection $\pi_f$ equals the group of translational motions. 
As a consequence a component of $V$, which corresponds to a pure translational motion, is projected to a single point $\go O$
(with $N\neq 0$) of the Euler parameter space $P^3$ by the elimination of $f_0,\ldots ,f_3$. 
Therefore the intersection of $\go O$ and $N=0$ equals $\varnothing$, which reasons the exclusion of pure translational motions 
within this approach. 

Moreover it is important to note that the set of bonds only depends on the geometry of the pentapod, 
and not on the leg lengths (cf.\ Theorem 1 of \cite{nawratil_bond}).

\begin{rmk} 
A more sophisticated bond theory for pentapods and hexapods is based on a special compactification of SE(3), where the 
sphere condition $\Lambda_i$ is only linear in 17 motion-parameters. This approach, which was presented in \cite{gns1}, 
has many theoretical advantages compared to the method described above, but it is not 
suited for direct computations due to the large number of motion-parameters. 
In contrast the approach of the paper at hand was already successfully used for direct computations 
in  \cite{nawratilmulti,nawratil_bond,nawratil_congruent,nawratil_equiform}. \hfill $\diamond$
\end{rmk}

Due to Theorem 2\footnote{This theorem is originally stated for hexapods but 
it also holds for pentapods, as its proof is also valid for $5$-legged manipulators.}
of \cite{nawratil_bond} a pentapod possesses a pure translational self-motion, 
if and only if the platform can be rotated about the center $\go m_1=\go M_1$ into a pose, where the vectors $\overrightarrow{\go M_i\go m_i}$ for $i=2,\ldots ,5$ 
fulfill the condition  $rk(\overrightarrow{\go M_2\go m_2},\ldots ,\overrightarrow{\go M_5\go m_5})\leq 1$.
Moreover all 1-dimensional self-motions are circular translations, which can easily 
be seen by considering a normal projection of the manipulator in direction of the parallel vectors $\overrightarrow{\go M_i\go m_i}$ for $i=2,\ldots ,5$. 
If all these five vectors are zero-vectors, which 
corresponds with the case that the platform and the base are congruent, 
we get the already mentioned 2-dimensional translational self-motion of a pentapod. 
This finishes already the discussion of pentapods with pure translatoric self-motions.

\subsection{Classification of 2-dimensional self-motions}\label{sec:class}

We assume that a given pentapod has a 2-dimensional self-motion $\mathcal{S}$. 
As $\mathcal{S}$ corresponds with a 
2-dimensional solution of the direct kinematics problem, the corresponding algebraic motion is also 2-dimensional. 
Due to the fact that this algebraic motion is the kinematic image of a self-motion, it cannot be located within the exceptional cone $N=0$. 
Therefore the bond-set $\mathcal{B}$ of this self-motion has to be an algebraic variety of dimension $1$; i.e.\ a {\it bonding curve}.

Now we want to classify $\mathcal{S}$ with respect to the dimension $\beta$ of $\mathcal{B}_f$. 
This classification was already used successfully for the determination of all hexapods with $(n>2)$-dimensional 
self-motions (cf.\  \cite{nawratilmulti}). 

As we have a bonding curve in the Study parameter space $P^7$, $\beta$ can take the values $-1,0,1$, 
where the case $\beta =1$ is the general one.  
In order that $\beta =i$ holds for $i=-1,0$, there has to exist a $(1-i)$-dimensional translational sub-self-motion, which is contained in $\mathcal{S}$, 
in each pose of $\mathcal{S}$. 
For $i=-1$ this already implies that $\mathcal{S}$ is a 2-dimensional translation; i.e.\ that platform and base are congruent. 
As $\beta=-1$ is already known, we only have to check the cases $\beta =1$ and $\beta =0$. 

\begin{rmk}\label{rem:1}
Note that the self-motions given in Theorem \ref{thm2} and item 1 of Theorem \ref{thm3} are of type $\beta =-1$. 
We will see within the proof of Theorem \ref{thm3} that the self-motion of item 2 is of type $\beta=0$ (see also Example 2 of \cite{nawratilmulti}).   
All designs of Theorem \ref{thm4} are of type $\beta=1$ as the 2-dimensional self-motions are spherical. 
\hfill $\diamond$ 
\end{rmk}


\section{Proving Theorem \ref{thm2}}\label{sec:pthm2}

Due to the above given considerations we only have to discuss the cases $\beta=0$ and $\beta=1$. 

\subsection{$\beta=0$} 

As there exists a 1-dimensional translational sub-self-motion in each pose of the 2-dimensional self-motion, 
we can apply Theorem 2 of \cite{nawratil_bond} (cf.\ footnote 4). As a consequence the platform and the base of the pentapod 
have to be related by a congruence transformation, which can be: 

\subsubsection{Orientation preserving ($\Rightarrow$ congruent pentapod) }

Due to Lemma 1 of \cite{nawratil_congruent} (cf.\ footnote 4)  non-planar congruent pentapods cannot have  a 1-dimensional translational self-motion. 
Therefore the case $\beta=0$ cannot exist.

\subsubsection{Orientation reversing ($\Rightarrow$ reflection-congruent pentapod)}\label{ref:2par}

In this case we can apply Theorem 2 of \cite{nawratil_equiform} (cf.\ footnote 4). 
Therefore non-planar reflection-congruent pentapods possess a 
2-parametric set of orientations with 1-dimensional translational self-motions. 	
In the following we show by means of computation that $\mathcal{B}_f=\varnothing$ holds 
for these manipulators, which implies the non-existence of 2-dimensional self-motion of type $\beta=0$: 

Without loss of generality (w.l.o.g.) we can assume that the first four anchor points span a tetrahedron. Moreover 
we can  choose special coordinate systems in the platform and the base in a way that we get:
\begin{align*}
\Vkt m_1=\Vkt M_1&=(0,0,0)^T, &\quad \Vkt M_j&=(a_j,b_j,c_j)^T, \\ 
\Vkt m_2=\Vkt M_2&=(a_2,0,0)^T, &\quad \Vkt m_j&=(a_j,b_j,-c_j)^T, \\
\Vkt m_3=\Vkt M_3&=(a_3,b_3,0)^T, &\quad &\phm
\end{align*} 
with $b_3c_4\neq 0$ and $j=4,5$. 
In addition we can eliminate the factor of similarity by setting $a_2=1$. 
For this setup, it can easily be seen that the 2-parametric set of platform orientations, which 
cause 1-dimensional translational sub-self-motions, is determined by $e_3=0$. 
We have to distinguish the following cases:
		\begin{enumerate}[1.]
		\item
		$e_2\neq 0$: Under this assumption we can solve
		$\Psi,\Delta_{2,1}$ for $f_2,f_3$ w.l.o.g., where $\Delta_{i,j}:=\Lambda_i-\Lambda_j$ is only linear in $f_0,\ldots ,f_3$. 
		We plug the obtained solutions into 
		$\Delta_{3,1}$, $\Delta_{4,1}$ and $\Delta_{5,1}$. 
		The numerators of the resulting expressions are denoted by $G_3$, $G_4$ and $G_5$, respectively, 
		which are homogeneous cubic polynomials in $e_0,e_1,e_2$. Note that $G_3$, $G_4$ and $G_5$ do not depend on $f_0$ and $f_1$.  
		These two Study parameters only appear in $\Lambda_1$, but this equation is not of interest for the further computation of bonds. 
		
		We eliminate $e_0$ from $G_i$ by calculating the resultant $H_i$ of $G_i$ and $N$ with respect to $e_0$ for $i=3,4$.  
		Now $H_3$ can only vanish without contradiction (w.c.) for either $e_1 = \frac{a_3}{b_3}e_2$ or $e_1 = \frac{a_3-1}{b_3}e_2$. 
		In both cases $H_4$ has to be fulfilled identically. The resulting condition can in both cases be solved for $a_4$ w.l.o.g., but it 
		can easily be seen that none of the obtained solutions is real for $b_3c_4\neq 0$, which finishes this case. 
 		\item
		$e_2= 0$:
		Now we can assume that $e_1\neq 0$ holds, as otherwise the orientation of the platform is fixed under the 2-dimensional 
		self-motion ($\Rightarrow$ pure translational motion). 
		Under this assumption we can solve $\Psi,\Delta_{3,1}$ for $f_1,f_3$ w.l.o.g.\ and plug the obtained solutions into 
		$\Delta_{2,1}$, $\Delta_{4,1}$ and $\Delta_{5,1}$. The numerators of the resulting expressions are denoted by $G_2$, $G_4$ and $G_5$, respectively. 
		$G_2$ equals $N(R_1^2-R_2^2)$ and $G_4,G_5$ are homogeneous cubic polynomials in $e_0,e_1$. 
		
		We eliminate $e_0$ from $G_4$ by calculating the resultant of $G_4$ and $N$ with respect to $e_0$ which yields:
		\begin{equation*}
		16b_3^2e_1^6(b_4^2+c_4^2)\left[(b_3-b_4)^2+c_4^2\right].
		\end{equation*}
		This expression cannot vanish w.c.\ over $\RR$ for $b_3c_4\neq 0$.
  	\end{enumerate}

\subsection{$\beta=1$}

We attach a sixth leg with anchor points $\go m_6$ and $\go M_6$ in a way that the platform and the base of the resulting
hexapod are still similar. This is a so-called equiform hexapod. 
If the pentapod has a 2-dimensional self-motions of type $\beta=1$, the equiform hexapod has to have a non-empty set $\mathcal{B}_f$, 
which is defined analogously to the one of pentapods (cf.\ Definition \ref{df:bond5} and Eq.\ (\ref{def:projbond})).

Due to \cite{nawratil_equiform} (and \cite{nawratil_congruent} for the special case of congruence) the following necessary condition 
for the existence of a projected bond $\in\mathcal{B}_f$ of an equiform (congruent) hexapod is known: 
The anchor points have to be located on a cylinder of revolution over $\CC$. 

Therefore the points $\go m_1,\ldots ,\go m_6$ of the resulting equiform hexapod have to be located on 
a cylinder of revolution $\Phi$ for any choice of $\go m_6$. The special choice $\go m_6\in [\go m_i,\go m_j]$ 
implies that the line $[\go m_i,\go m_j]$ has to be contained in $\Phi$. 
Therefore $\Phi$ can only be (cf.\ \cite{nawratil_congruent} or \cite{nawratil_equiform}):
\begin{enumerate}[$\star$]
\item
either a cylinder of revolution over $\RR$ and $[\go m_i,\go m_j]$ is one of its rulings,
\item
or it splits up into a pair of isotropic planes, which are not conjugate complex.
In this case $\Phi$ carries two real lines.
\end{enumerate}

Therefore the following statement has to \medskip hold:  

\noindent
{\it Given are 5 real points $\go m_1,..,\go m_5$, which are not coplanar. 
For each of the 10 lines $[\go m_i,\go m_j]$ $i\neq j\in\left\{1,2,3,4,5\right\}$
one of the following two properties has to hold: 
\begin{enumerate}
\item
$\go m_1,..,\go m_5$ are located on a real cylinder of revolution with the ruling $[\go m_i,\go m_j]$.
\item
There exists a second line skew to $[\go m_i,\go m_j]$ in a way that $\go m_1,..,\go m_5$ are located on both lines.
\end{enumerate}
}

In the following we show that the second possibility yields a contradiction:
If  $\go m_1,..,\go m_5$ are located on two skew lines, then we can assume that no 4 points are collinear, as otherwise 
we contradict our non-planarity assumption. W.l.o.g.\ we can assume that $\go m_1,\go m_2,\go m_3$ are located on the line $\go g$. 
If we consider for example the line $[\go m_1,\go m_4]$, it can easily be seen that no cylinder of revolution $\Phi$ with the ruling $[\go m_1,\go m_4]$ 
passing through $\go m_2,\go m_3,\go m_5$ can exist, as $\go g$ can intersect $\Phi$ only in two points. 
Therefore we can reduce the problem to the following \medskip question: 

\noindent
{\it Given are 5 real points $\go m_1,..,\go m_5$, which are not coplanar and where no three points are collinear. 
Can a configuration of these 5 points exist in a way that they are located on the 
10 real cylinders of revolution with rulings $[\go m_i,\go m_j]$ for \medskip $i\neq j\in\left\{1,2,3,4,5\right\}$?}

The answer is "no", which can be proven as follows: 
There has to exist a plane $\varepsilon_{ij}$ containing $[\go m_i,\go m_j]$ with the property that the remaining 
three points are located in the same half-space with respect to $\varepsilon_{ij}$ but do not belong to $\varepsilon_{ij}$. 
The reason for this is that $\varepsilon_{ij}$ can be seen as tangent plane to the cylinder along the ruling $[\go m_i,\go m_j]$. 

Therefore the segment $\overline{\go m_i\go m_j}$ has to be an edge of the convex hull of the point set $\go m_1,..,\go m_5$. 
As a consequence no 4 points of $\go m_1,..,\go m_5$ can be coplanar. 
Therefore the convex hull has to be a convex polyhedron with 5 vertices, 10 edges and 10 faces. As this
contradicts Euler's polyhedron formula $V-E+F=2$,  Theorem \ref{thm2} is proven.  \hfill $\BewEnde$


\section{Proving Theorem \ref{thm3}}\label{sec:pthm3}

Due to the considerations given in Section \ref{sec:class} we only have to discuss the cases $\beta=0$ and $\beta=1$.

\subsection{$\beta=1$} \label{sec:b1}

Assume that a planar affine pentapod with a 2-dimensional self-motion of type $\beta=1$ is given. 
Now we attach a sixth leg in a way that we get a planar affine hexapod, which is not 
architecturally singular.\footnote{This can always be done if the given pentapod is not architecturally singular.} 
Therefore the resulting hexapod has to have a non-empty set $\mathcal{B}_f$. 
In Example 5 of \cite{nawratil_bond} it is proven that a planar affine hexapod does not possess a projected bond. 
Therefore a planar affine pentapod cannot have a 2-dimensional self-motions of type $\beta=1$.

\subsection{$\beta=0$} 

This case has to be subdivided with respect to the criterion if the affinity $\alpha$ is a congruence transform or not.

\subsubsection{The affinity is no congruence transformation}\label{aff:general}

We assume that a pentapod with a 2-dimensional self-motion $\mathcal{S}$ of type $\beta=0$ is given. Therefore it 
can perform in each configuration $\mathcal{C}$ of $\mathcal{S}$ a 1-dimensional translational self-motion $\mathcal{T}$. 
If we disconnect the platform and the base in the pose $\mathcal{C}$ and translate the platform in a way that $\go M_1=\go m_1$ holds, then 
the following relation has to be fulfilled due to the last paragraph of Section \ref{sec:def}: 
\begin{equation*}
rk(\overrightarrow{\go M_2\go m_2},\ldots ,\overrightarrow{\go M_5\go m_5})= 1.
\end{equation*}
Now the following two cases have to be distinguished: 
\begin{enumerate}[1.]
\item
{\bf For all $\mathcal{C}$ of $\mathcal{S}$ the platform is parallel to the base:} 
Therefore $\mathcal{S}$ has to be a Sch\"onflies motion where the direction of the rotational axis is orthogonal to the base plane. 
In the following we want to compute the affine pentapods with this property: 

W.l.o.g.\ we can assume that $\go M_1$ equals in the origin of the fixed system, that $\go M_2$ is located on its $x$-axis and that 
the remaining base anchor points belong to the $xy$-plane. The same can be assumed for the platform with respect to the moving system. 
Moreover we can assume $\go m_1,\go m_2,\go m_3$ as well as $\go M_1,\go M_2,\go M_3$ are not collinear (otherwise we can relabel the anchor points). 
Therefore the (regular) affinity between the platform and base is determined by the first three pairs of anchor points with coordinates:
\begin{equation*}
\Vkt M_1=\Vkt m_1=(0,0,0)^T,\quad \Vkt m_2=(a_2,0,0)^T,\quad \Vkt m_3=(a_3,b_3,0)^T,\quad \Vkt M_2=(A_2,0,0)^T,\quad \Vkt M_3=(A_3,B_3,0)^T,
\end{equation*} 
and $a_2A_2b_3B_3\neq 0$.
Now we rotate the platform about the $z$-axis which yields $\Vkt m_i^{\prime}:=\Vkt R\Vkt m_i$ where $e_1$ and $e_2$ have to be set equal to zero in 
the matrix $\Vkt R$ of Eq.\ (\ref{mat:R}). Now the necessary condition for the existence of a translational self-motion reads as:
\begin{equation*}
(\Vkt m_2^{\prime}-N\Vkt M_2)\times (\Vkt m_3^{\prime}-N\Vkt M_3)=\Vkt o.
\end{equation*}
This only implies the following equation:
\begin{equation*}
(e_0^2+e_3^2)\left[
(B_3-b_3)(A_2-a_2)e_0^2+
(B_3+b_3)(A_2+a_2)e_3^2+
2(A_3a_2-A_2a_3)e_0e_3
\right].
\end{equation*}
The second factor is fulfilled identically if and only if the 
platform and the base are congruent, which is the excluded case. 
Therefore there exist at most two orientations which cause a translational self-motion $\mathcal{T}$, but for 
a 2-dimensional self-motion $\mathcal{S}$ of type $\beta=0$ we need at least a 1-dimensional set of such orientations. 
Hence we have no solution in this case.
\begin{figure}[top]
\begin{center}  
 \begin{overpic}  
    [width=43mm]{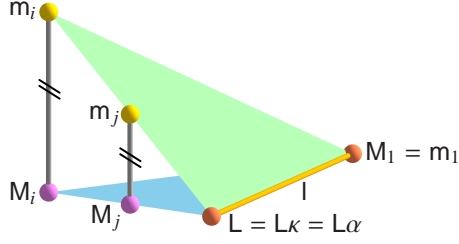}
	\put(100.5,22.5){$\go M_1=\go m_1$}
	\put(-8.5,67){$\go m_i$}
	\put(16,35){$\go m_j$}	
	\put(58.5,0){$\go L=\go L\kappa=\go L\alpha$}
	\put(-8.5,10){$\go M_i$}
	\put(16.5,3){$\go M_j$}
	\put(81,10){$\go l$}
  \end{overpic} 
\end{center}
\caption{The parallel projection $\kappa$ from the base plane to the non-coinciding 
platform plane equals the affinity $\alpha$.}
  \label{figextra} 
\end{figure}  
\item
{\bf There exists a pose $\mathcal{C}$ of $\mathcal{S}$ where the platform is not parallel to the base:} 
After performing the translation of the platform that  $\go M_1=\go m_1$ holds there 
exists a parallel projection $\kappa$ which maps $\go M_i$  to $\go m_i$ for $i=1,\ldots, 5$ (see Fig.\ \ref{figextra}). 
Therefore this mapping equals the affine mapping $\alpha$ between the base and the platform. 

As a consequence the points $\go L$ on the line $\go l$ of intersection of the platform and the base have to be mapped onto each other. 
It is a well known fact from linear algebra that there exist at most two distinct directions $\Vkt d_1,\Vkt d_2$,
which are mapped without distortion (i.e.\ $\|\Vkt d_i\|=\|\alpha(\Vkt d_i)\|$ for $i=1,2$), if $\alpha$ is no congruence transformation. 
Therefore the line $\go l$ has to be parallel to $\Vkt d_1$ and $\alpha(\Vkt d_1)$, which are equally oriented. 
Moreover the orientations with translational self-motions are obtained by rotating the platform about the line $\go l$. 
Within this 1-parametric set of orientations there are two poses where the platform and the base coincide. 
It is impossible that in one of these two flat configurations also $\Vkt d_2$ and $\alpha(\Vkt d_2)$ are parallel and equally oriented, 
as otherwise the affinity has to be congruence transformation. Therefore no bifurcation into another 1-parametric set of orientations 
causing translational self-motions is possible.
Therefore $\mathcal{S}$ has to be a Sch\"onflies motion where the direction of the rotational axis equals $\Vkt d_1$.

With the obtained information the remaining problem can easily be solved by direct computations as follows: 
W.l.o.g.\ we can choose the fixed system that $\go M_1$ equals its origin and that the $x$-axis shows in direction $\Vkt d_1$. 
Moreover, $\go m_1$ is the origin of the moving system and its $x$-axis corresponds with the direction $\alpha(\Vkt d_1)$. 
Therefore the anchor points have the following coordinates:
\begin{equation*}
\Vkt M_1=\Vkt m_1=(0,0,0)^T, \quad  \Vkt M_j=(A_j,B_j,0)^T,\quad \Vkt m_j=(A_j,kB_j,0)^T,
\end{equation*} 
for $j=2,\ldots ,5$. Due to the planarity of the platform we can assume $k\in\RR^+$, where the value $0$ has to 
be excluded\footnote{One can also exclude $k=1$ as it equals the congruent case, but the following 
calculation also holds for this case.} as otherwise $\alpha$ is singular. 
Moreover we know that $\mathcal{S}$ is a Sch\"onflies motion with respect to $\Vkt d_1$ and 
therefore the Euler parameters $e_2$ and $e_3$ have to be equal to zero. 

According to Remark \ref{rmk1} a translation of a leg along $\Vkt d_1$ does not influence the Sch\"onflies motion. 
Therefore we can set $A_j=0$ for $j=2,\ldots ,5$ w.l.o.g.. As not all base anchor points are collinear we can also assume that $B_2\neq 0$ holds.
Under this assumption we can solve $\Psi,\Delta_{2,1}$ for $f_1,f_3$. 
Then we plug the obtained solutions into $\Delta_{i,1}$ with $i=3,4,5$. The numerators of the resulting expressions are denoted by $G_i$ 
and they are homogeneous quadratic in $e_0,e_1$. Note that $G_i$ does not depend on $f_0$ and $f_2$. These two unknowns only remain\footnote{It can easily 
be verified that it is impossible that $\Lambda_1$ is also independent of  $f_0$ and $f_2$.} in $\Lambda_1$. 

A projected bond exists if $G_3=0,G_4=0,G_5=0$ and $e_0^2+e_1^2=0$ ($\Leftrightarrow$ $N=0$) have a common root.
Hence we compute the resultant of $G_i$ and $N$ with respect to $e_1$ which yields 
\begin{equation*}
16B_2^2B_i^2k^2e_0^4(B_2-B_i)^2.
\end{equation*}
As a consequence the $i$th leg has to coincide with the first or second leg for $i=3,4,5$. 
As the collinearity of four anchor points yields a trivial case of an architecturally singular design (cf.\ item 8 of Theorem 3 given by Karger \cite{kargernonplanar}), 
we end up with the manipulator given in item 2 of Theorem \ref{thm3}. This manipulator design is sufficient for the existence 
of a 2-dimensional self-motion $\mathcal{S}$ due to Remark \ref{rmk1}. 
\end{enumerate}


\subsubsection{The affinity is a congruence transformation}\label {aff:spezi}

In this case we can assume coordinate systems in the platform and the base in a way that the points are located in the $xy$-plane and that 
$\Vkt M_i=\Vkt m_i$ holds for $i=1,\ldots ,5$. As in Section \ref{ref:2par} it can easily be seen that the set of orientations causing 
translational self-motions is 2-dimensional given by the condition $e_3=0$ in the Euler parameters. 
Moreover due to the result obtained in item 2 of Section \ref{aff:general} (cf.\ footnote 6) we can assume that the ratio
$e_1:e_2$ is not constant during the 2-dimensional self-motion $\mathcal{S}$ as otherwise $\mathcal{S}$ has to be a Sch\"onflies motion, 
which can only result in a special case of the solution given in item 2 of Theorem \ref{thm3}. 

Now we check the remaining cases by means of computer algebra. 
To do this in a clear way, it is split up with respect to criterion whether  three collinear anchor points exist or not. 

\begin{rmk}
If no three anchor points are collinear then the pentapod is free of so-called butterfly self-motions (cf.\ \cite{gns1,nawratil_bond}). 
Therefore the splitting with respect to the existence/non-existence of three collinear anchor points is equivalent with the 
splitting with respect to the existence/non-existence of butterfly self-motions. \hfill $\diamond$
\end{rmk}

\begin{enumerate}[1.]
\item
{\bf No three anchor points are collinear:} 
It is well known (cf.\ Bricard \cite{bricard}, Chasles \cite{chasles}, Duporcq \cite{duporcq2}) 
that one can add a 1-parametric set of legs to a planar projective pentapod without changing the direct kinematics. 
The anchor points of these legs are also related by the projectivity and they are located on the conic sections uniquely 
determined by the given five pairs of anchor points (see Fig.\ \ref{fig05}a). 
As a congruence $\alpha$ is only a special case of a projectivity, we can use this result for the problem under consideration. We denote the 
conic section in the base by $\go c$. Hence the corresponding one in the platform is given by $\go c\alpha$. 
Due to our assumption of the non-collinearity of three points, the conic $\go c$ has to be regular; i.e.
an ellipse, a hyperbola or a parabola. 

\begin{figure}[top] 
\centering
\subfigure[]{
 \begin{overpic}
    [width=32mm]{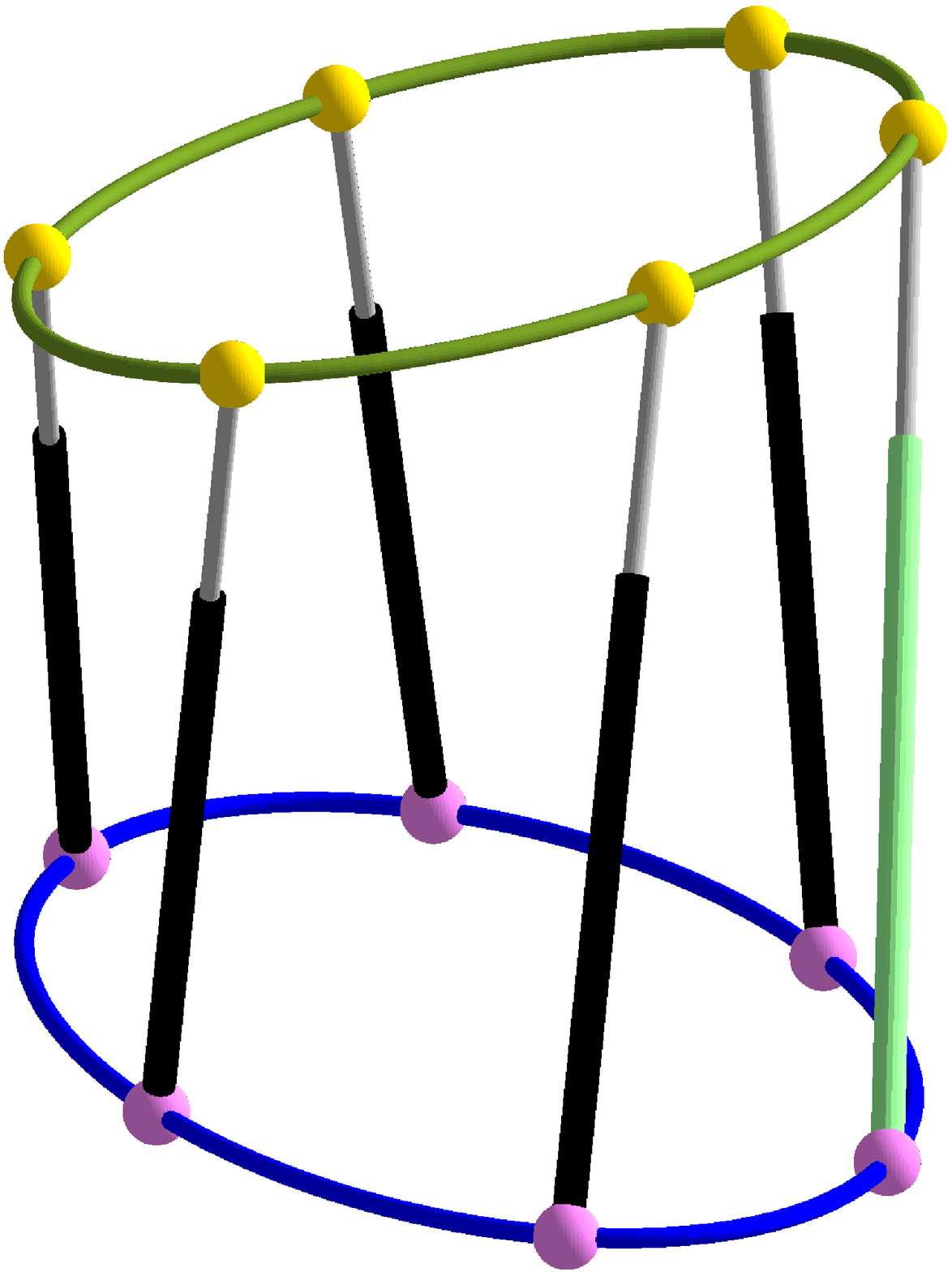}
	\begin{small}
    \put(25,9){\contour{white}{$\go c$}}    
 		\put(41,89){\contour{white}{$\go c\alpha$}}
	\end{small}

  \end{overpic}
  }
  \quad 
\subfigure[]{
\begin{overpic}
    [width=37.5mm]{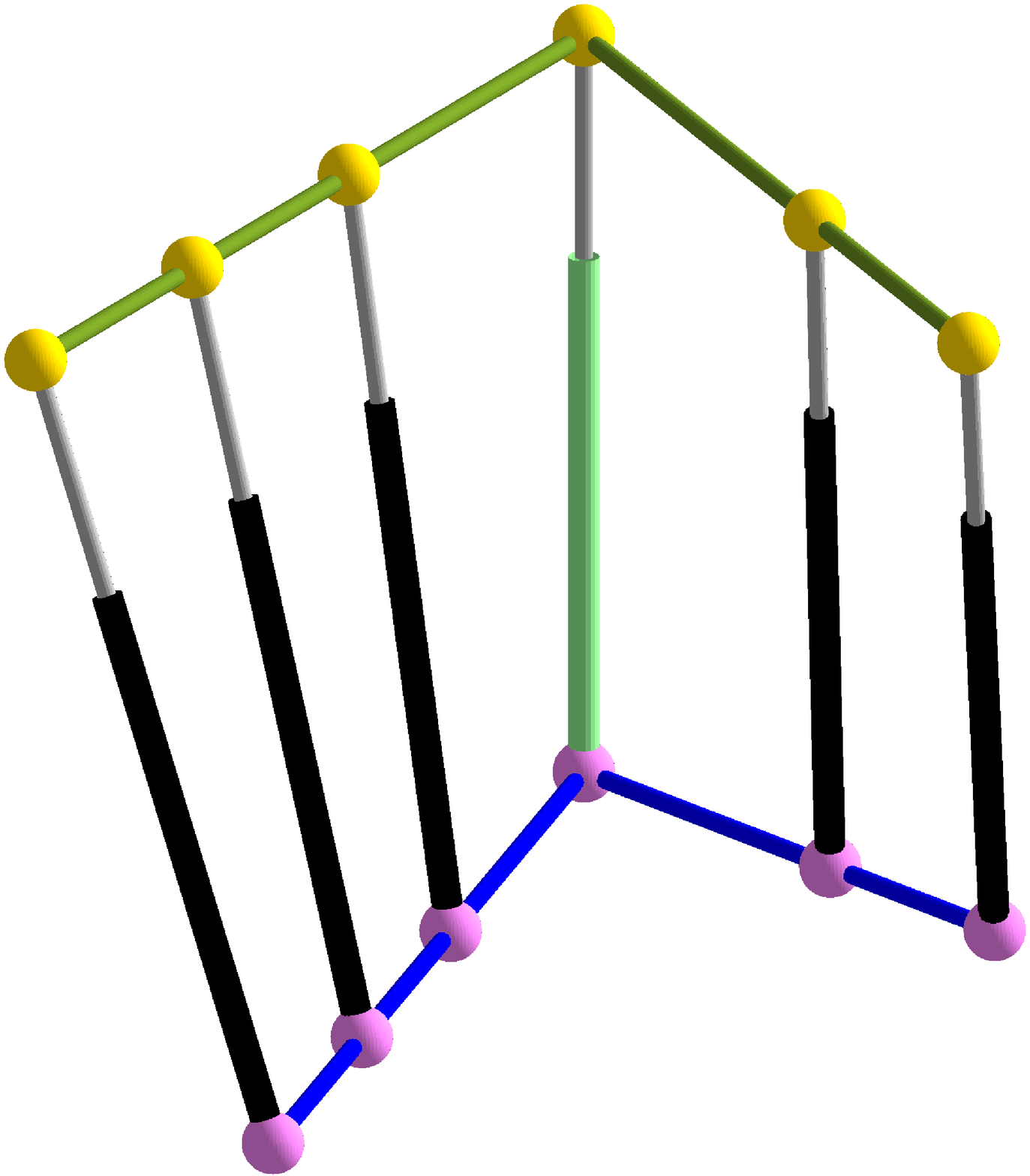}
	\begin{small}
    \put(56,24.5){\contour{white}{$\go c$}}    
 		\put(33,93){\contour{white}{$\go c\alpha$}}
	\end{small}
  \end{overpic}
}
\caption{A 1-dimensional set of legs can be attached  to a planar projective pentapod. Only one additional 
leg of this set is visualized in green color: (a) general case (b) conic $\go c$ splits up into two lines. 
}  
\label{fig05}
\end{figure}

W.l.o.g.\ we can choose the coordinate system in the base in a way that the origin coincides with a finite point $\go M_1\in\go c$. 
Then we can still rotate the coordinate system about this point by any angle $\delta\in[0,2\pi)$. 
Now we consider the following four lines with respect to the fixed system: 
\begin{equation*}
y=0,\quad x=0,\quad x=y,\quad x=-y.
\end{equation*}
We assume that $\delta$ is chosen in a way that none of these four lines equals the tangent in $\go M_1$ with respect to $\go c$ 
or intersects $\go c$ in an ideal point. This assumption can be done w.l.o.g.\ as there can only exist a finite number of such "bad" choices for 
$\delta$. 
Therefore we get the following coordinatization:
\begin{align*}
\Vkt M_1&=(0,0,0)^T, &\quad \Vkt M_2&=(A_2,0,0)^T, &\quad \Vkt M_3&=(0,B_3,0)^T, \\
\Vkt M_4&=(A_4,A_4,0)^T, &\quad \Vkt M_5&=(A_5,-A_5,0)^T, &\quad &\phm
\end{align*}
with  $A_2B_3A_4A_5\neq 0$. Moreover we can assume w.l.o.g.\ that $A_2>0$ hold; i.e.\ $\go M_2$ is located on the 
positive $x$-axis of the fixed frame. 
The moving frame is chosen analogously thus we get $\Vkt m_i=\Vkt M_i$ for $i=1,\ldots ,5$.

We solve $\Psi,\Delta_{2,1}$ for $f_2,f_3$ and plug the obtained solutions into $\Delta_{3,1},\Delta_{4,1},\Delta_{5,1}$. 
The numerators of the resulting expressions are denoted by $G_3,G_4$ and $G_5$, respectively, and they are 
homogeneous cubic polynomials in $e_0,e_1,e_2$. Note that $G_3,G_4,G_5$  
do not depend on $f_0$ and $f_1$. These two unknowns only remain\footnote{It can easily 
be verified that it is impossible that $\Lambda_1$ is also independent of  $f_0$ and $f_1$.} in $\Lambda_1$. 

Now a projected bond exists if the three cubic curves $G_3=0,G_4=0,G_5=0$ have a common 
intersection point with the conic $e_0^2+e_1^2+e_2^2=0$ ($\Leftrightarrow$ $N=0$). 
In the following we show that this is not possible: 

We compute the resultant $H_i$ of $G_i$ and $N$ with respect to $e_0$. For 
$H_3$ we get $16B_3^2e_1^2e_2^2(B_3e_1+e_2)^2$. As $e_2$ is the denominator of $f_2$ and $f_3$ we have to distinguish the following cases: 
	\begin{enumerate}[(a)]
	\item
	$e_2\neq 0$: We remain with two possibilities:
		\begin{enumerate}[i.]
		\item
		$e_1=0$: Then $H_j$ equals $16e_2^6A_j^2(A_j-1)^2$ for $j=4,5$. 
		This implies $A_4=A_5=1$, but in this case $\go M_2,\go M_4,\go M_5$ are collinear. 
		\item
		$e_1 = -e_2/B_3$: In this case we get
		\begin{equation*}
		H_4= 16A_4^2B_3^{-4}e_2^6(B_3+1)^2(B_3A_4-B_3+A_4)^2, \qquad
		H_5= 16A_5^2B_3^{-4}e_2^6(B_3-1)^2(B_3A_5-B_3+A_5)^2.
		\end{equation*}
		As the last factor yields the collinearity of $\go M_2,\go M_3,\go M_4$ and $\go M_2,\go M_3,\go M_5$, respectively, 
		we are done. 
		\end{enumerate}
	\item
	$e_2=0$: Finally we have to show that $(1:\pm i:0:0)$ is no projected bond of our manipulator. Under consideration of these Euler parameters we can 
	solve $\Psi,\Delta_{3,1}$ for $f_1,f_3$ which yields
	$f_1=\pm if_0$ and $f_3=\mp iB_3/2$, respectively. 
	Then $\Delta_{4,1}$ and $\Delta_{5,1}$ can only vanish w.c.\ for $B_3=A_4=-A_5$, but in this case $\go M_3,\go M_4,\go M_5$ are collinear. 
	\end{enumerate}

\item
{\bf Three anchor points are collinear:} 
Due to this collinearity the 
conic section $\go c$ splits up into two lines (see Fig.\ \ref{fig05}b). 
Moreover we can assume that no four anchor points are collinear, as otherwise we get an 
architecturally singular pentapod. 

W.l.o.g.\ we can assume that $\go M_1,\go M_2,\go M_3$ are located on the $x$-axis. 
Due to the result of  Bricard \cite{bricard}, Chasles \cite{chasles} and Duporcq \cite{duporcq2}, 
we can even fix their coordinates as follows:
\begin{equation*}
\Vkt M_1=(0,0,0)^T,\quad 
\Vkt M_2=(1,0,0)^T,\quad 
\Vkt M_3=(-1,0,0)^T.
\end{equation*}

	\begin{enumerate}[(a)]
	\item
	$[\go M_4,\go M_5]$ is not parallel to the $x$-axis:
	In this case we can assume w.l.o.g.\ that their point of intersection is the origin (= $\go M_1$). Now the remaining two 
	base anchor points can be coordinatized as: 
	\begin{equation*}
	\Vkt M_4=(A_4,1,0)^T,\quad 
	\Vkt M_5=(-A_4,-1,0)^T.
	\end{equation*}
	The coordinates of the corresponding platform anchor points are determined by $\Vkt m_i=\Vkt M_i$ for $i=1,\ldots ,5$.

	We can solve $\Psi,\Delta_{2,3}$ w.l.o.g.\ for $f_2$ and $f_3$.
	Then we plug the obtained solutions into $\Delta_{2,1}$, $\Lambda_4+\Lambda_5$ and $\Delta_{4,5}$. 
	The numerators of the resulting expressions are denoted by $G_2$, $G_4$ and $G_5$, respectively, 
	and they do not depend on $f_0$ and $f_1$. 
	$G_2$ and $G_4$ are homogeneous quadratic polynomials in $e_0,e_1,e_2$ and 
	$G_5$ factors into $(e_0^2+e_1^2+e_2^2)H_5$ with:
	\begin{equation*}
	H_5=(R_3^2-R_2^2)e_1 + \left[R_5^2-R_4^2-A_4(R_3^2-R_2^2)\right]e_2.
	\end{equation*}
	Therefore $H_5$ has to be fulfilled identically, as otherwise the ratio $e_1:e_2$ is constant. 
	This implies $R_2^2=R_3^2$ and $R_4^2=R_5^2$. 
	
	In order that a 2-dimensional  self-motion of type $\beta=0$ exists, the two remaining equations $G_2=0$ and $G_4=0$ have to have a common factor. 
		Therefore we compute the resultant of these two expressions with respect to $e_0$, which yields 
		$16L^2$ with 
		\begin{equation*}
		L=(R_1^2-R_3^2)e_1^2+ \left[A_4^2(R_1^2-R_3^2)-R_1^2+R_5^2\right]e_2^2 -2A_4(R_1^2-R_3^2)e_1e_2.
		\end{equation*} 
		The necessary condition $R_1^2=R_3^2$ already implies $G_2=4e_2^2$, which cannot vanish w.c..
	\item
	$[\go M_4,\go M_5]$ is parallel to the $x$-axis:
	In this case the remaining two platform anchor points can be coordinatized as follows: 
	\begin{equation*}
	\Vkt M_4=(0,B_4,0)^T,\quad 
	\Vkt M_5=(1,B_4,0)^T.
	\end{equation*} 
	Moreover we can eliminate the factor of scaling by setting $B_4=1$. 
	The coordinates of the corresponding platform anchor points are determined by $\Vkt m_i=\Vkt M_i$ for $i=1,\ldots ,5$.

	We can solve $\Psi,\Delta_{2,3}$ w.l.o.g.\ for $f_2$ and $f_3$.
	Then we plug the obtained solutions into $\Delta_{2,1}$ and $\Delta_{4,5}$. 
	The numerators of the resulting expressions are denoted by $G_2$ and $G_4$, respectively, 
	and they do not depend on $f_0$ and $f_1$. 
	$G_2$ and $G_4$ are homogeneous quadratic polynomials in $e_0,e_1,e_2$. 
 
	In order that a 2-dimensional  self-motion of type $\beta=0$ exists, 
	the two equations $G_2=0$ and $G_4=0$ have to have a common factor. 
	Therefore we compute the resultant of these two expressions with respect to $e_0$, which yields: 
	\begin{equation*}
	64e_2^2\left[(2R_1^2-R_2^2-R_3^2)e_1-(R_1^2-R_2^2+R_5^2-R_4^2)e_2\right]^2.
	\end{equation*} 
	The necessary condition $R_1^2=(R_2^2+R_3^2)/2$ already implies $G_2=4e_2^2$, which cannot vanish w.c..
	This finishes the proof of Theorem \ref{thm3}.  \hfill $\BewEnde$
\end{enumerate}
\end{enumerate}
 

\section{Proving Theorem \ref{thm4}}\label{sec:pthm4}

As already mentioned in the last paragraph of Section \ref{intro} the proof of this theorem splits up into two parts, which are 
discussed in the following subsections:

\subsection{Study of item (c) of Theorem \ref{thm1a}}\label{part(c)}

We only have to discuss the cases $p=2,3,4$, as $p<2$ yields architecturally singular designs 
and $p=5$ the excluded case of a linear platform. 
But within the following study we allow the collinearity of the five base anchor points, as in this way we get the following 
three additional cases as a side-result:
\begin{itemize}
\item[$(\alpha)$]
We get a special case of item 1 of Theorem \ref{thm4} if $\go M_1,\go M_2,\go M_3=\go M_4=\go M_5$ are collinear: The  
2-dimensional spherical self-motion is obtained if $\go M_3=\go M_4=\go M_5$ is located on the line spanned by $\go m_1$ and $\go m_2$. 
\item[$(\beta)$]
We get a special case of item 2(b) of Theorem \ref{thm4} if $\go M_1,\go M_2,\go M_3,\go M_4=\go M_5$ are collinear: The 
2-dimensional spherical self-motion is obtained if $\go M_4=\go M_5$ is located on the carrier line of the collinear points $\go m_1,\go m_2,\go m_3$. 	
\item[$(\gamma)$]
We get a special case of item 3 of Theorem \ref{thm4} if $\go M_1,\go M_2,\go M_3,\go M_4,\go M_5$ are collinear: The  
2-dimensional spherical self-motion is obtained if $\go M_5$ is located on the carrier line of the collinear points $\go m_1,\ldots,\go m_4$.
\end{itemize}
Until now these three special cases were not reported in the literature to the best knowledge of the authors. 
Hence, their existence necessitates a contemporary and accurate re-examination of the old results on this topic 
(cf.\ page 222 of Darboux \cite{koenigs}, pages 180ff.\ of Mannheim \cite{mannheim}, 
Duporcq \cite{duporcq}; see also Chapter III of Bricard \cite{bricard}), which also takes the coincidence of anchor points into account.
This is done by the authors in \cite{ns}.

\begin{figure}[top] 
\centering
\subfigure[]{
 \begin{overpic}
    [width=37mm]{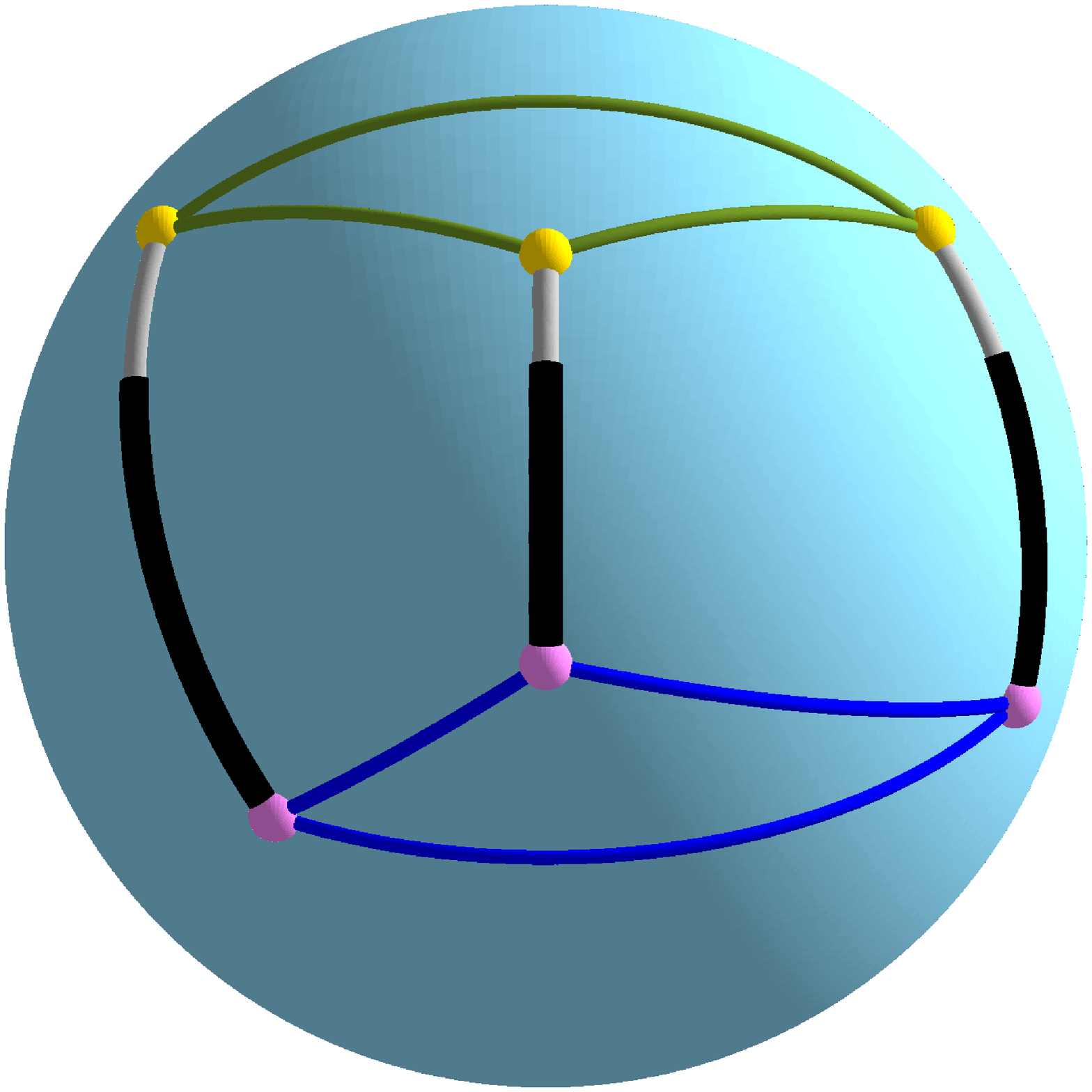}
  \contourlength{0.3mm}
  \begin{small} 
    \put(11.8,20.5){\contour{white}{$\go M_1^{\circ}$}}    
    \put(47.5,29){\contour{white}{$\go M_3^{\circ}$}}
    \put(80.5,40){\contour{white}{$\go M_2^{\circ}$}}
	\put(18,72){\contour{white}{$\go m_1^{\circ}$}}
	\put(40,82){\contour{white}{$\go m_3^{\circ}$}}
	\put(75,71){\contour{white}{$\go m_2^{\circ}$}}
	\end{small}
  \end{overpic}
  }
\subfigure[]{
\begin{overpic}
    [width=37mm]{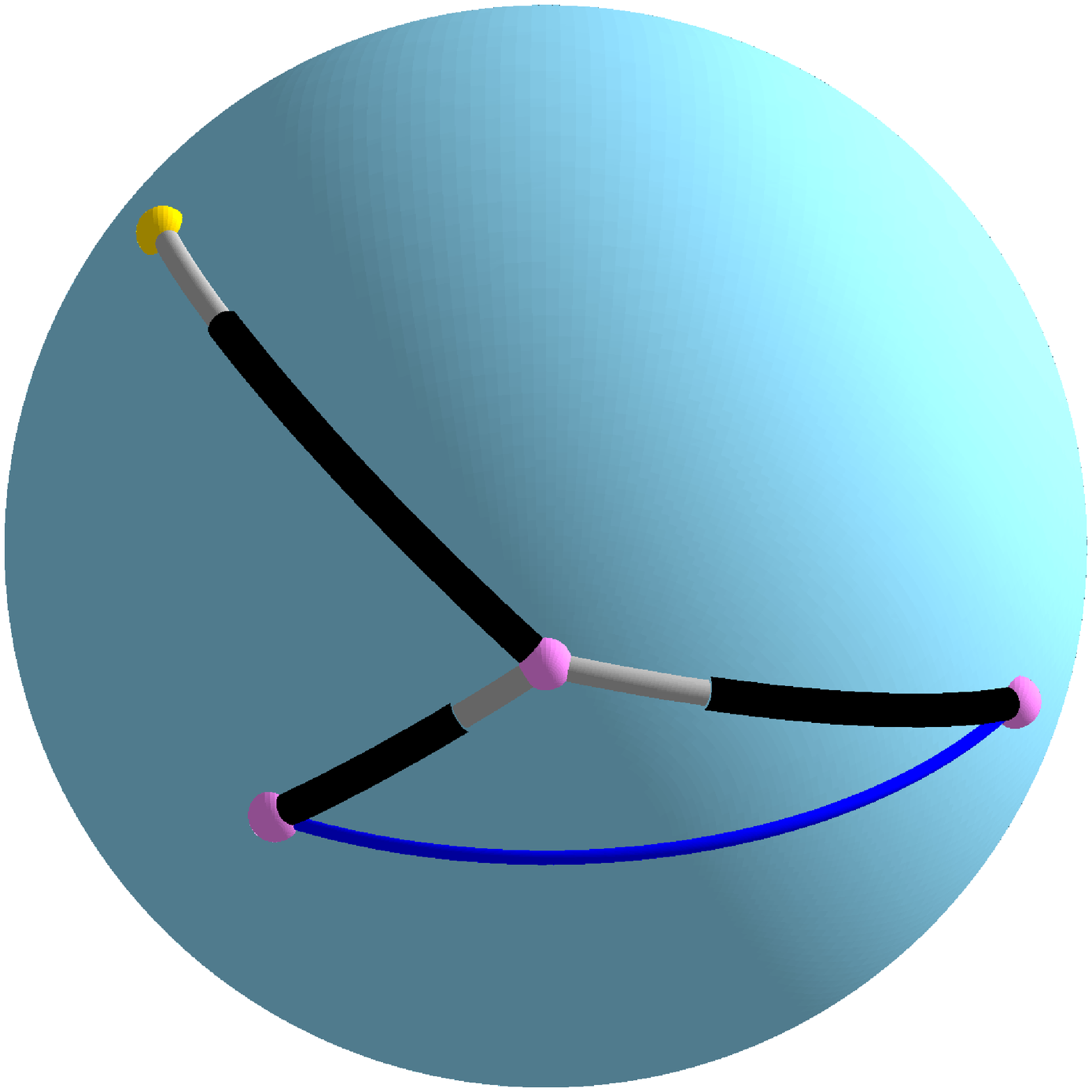}
    \contourlength{0.3mm}
    \begin{small}
    \put(12.5,24.5){\contour{white}{$\go M_1^{\circ}$}}    
    \put(47.5,29){\contour{white}{$\go M_3^{\circ}$}}
    \put(50,43.5){\contour{white}{$\go m_1^{\circ}= \go m_2^{\circ}$}}
    \put(88,41.5){\contour{white}{$\go M_2^{\circ}$}}
	\put(19,77.5){\contour{white}{$\go m_3^{\circ}$}}
	\end{small}
  \end{overpic}
}
\subfigure[]{
\begin{overpic}
    [width=37mm]{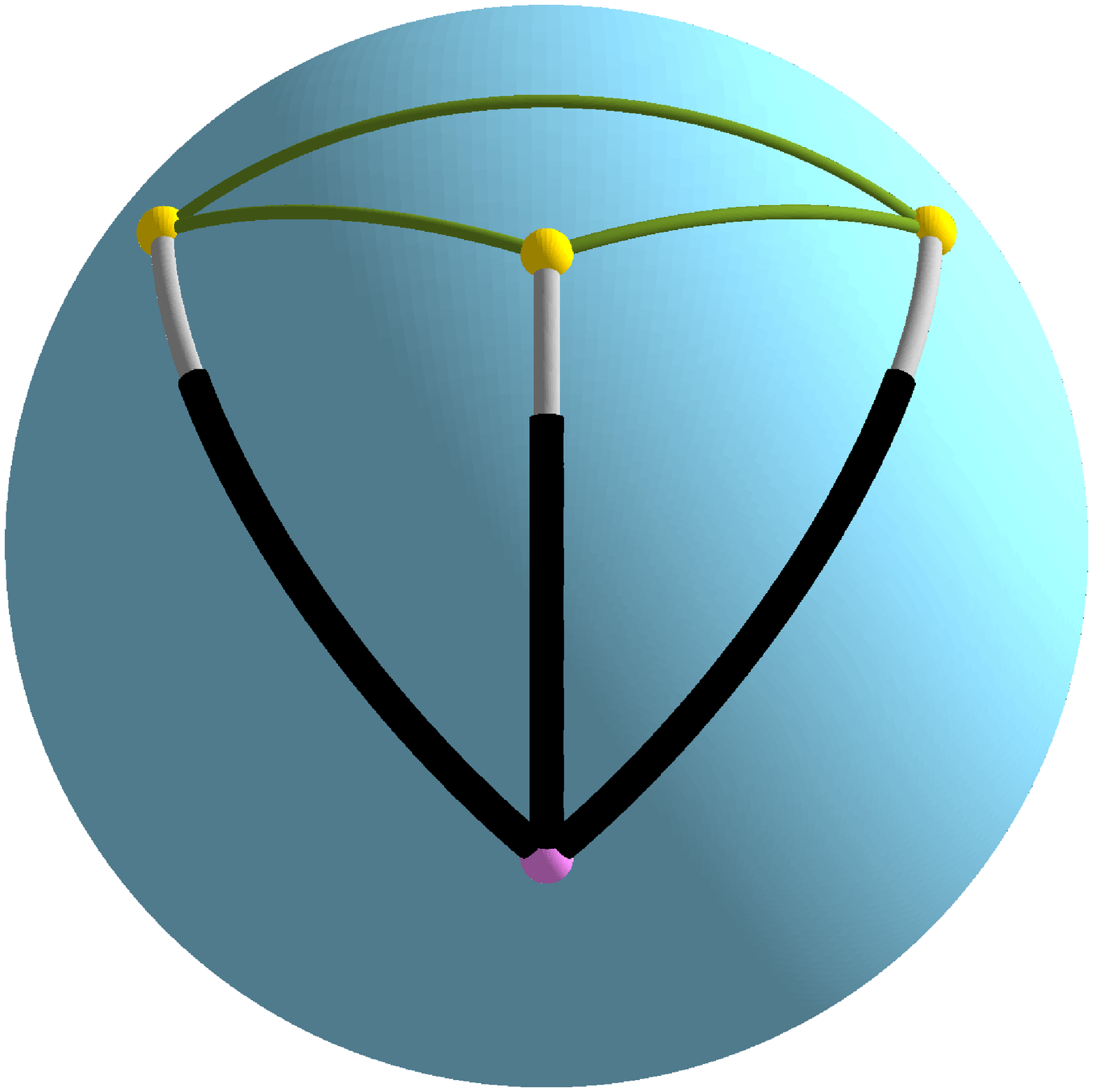}
    \contourlength{0.3mm}
    \begin{small}
    \put(24,11){\contour{white}{$\go M_1^{\circ}= \go M_2^{\circ}= \go M_3^{\circ}$}}
		\put(19,72){\contour{white}{$\go m_1^{\circ}$}}
		\put(40,82){\contour{white}{$\go m_3^{\circ}$}}
		\put(72,71.5){\contour{white}{$\go m_2^{\circ}$}}
	\end{small}
  \end{overpic}
}
\subfigure[]{
\begin{overpic}
    [width=37mm]{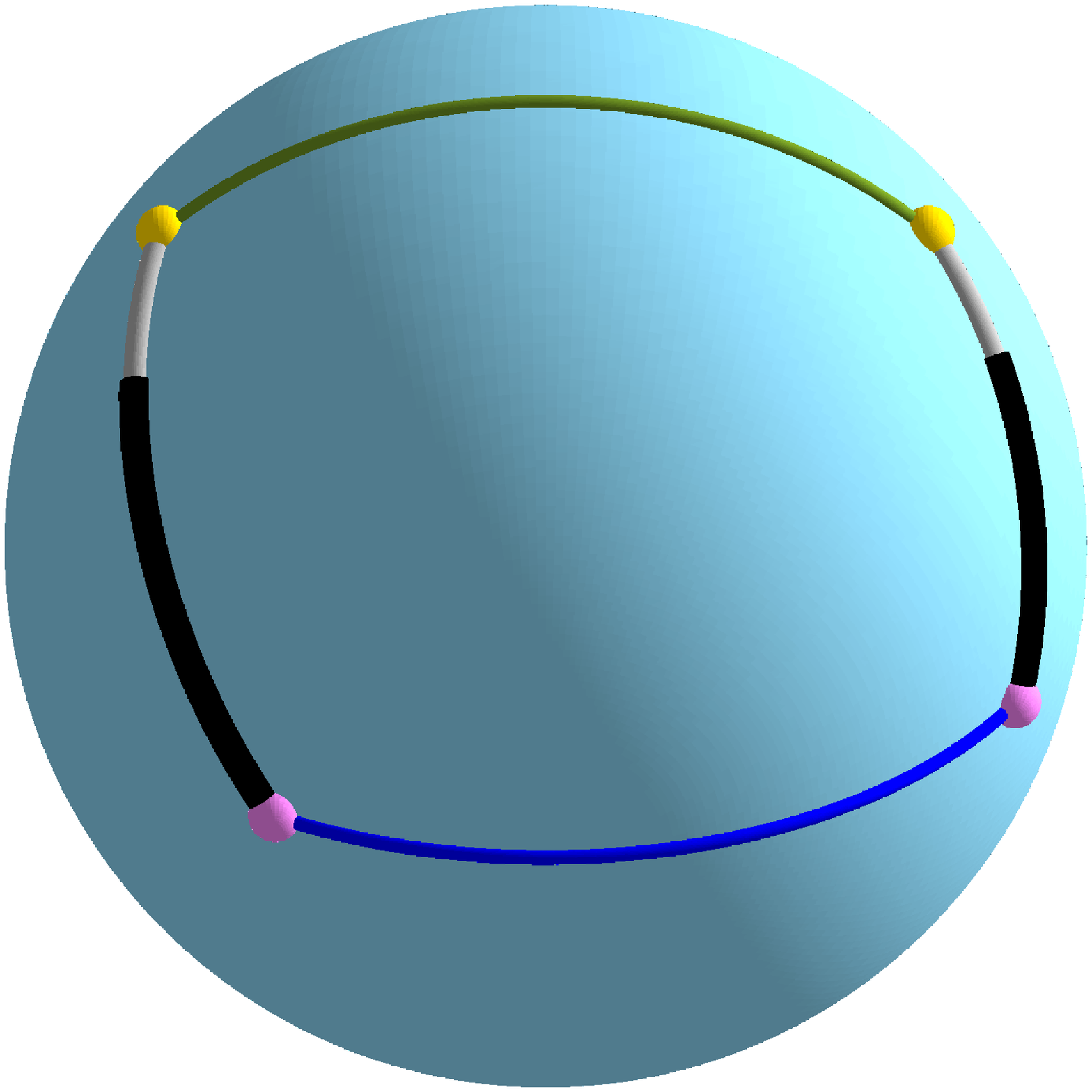}
    \contourlength{0.3mm}
    \begin{small}
    \put(28,28){\contour{white}{$\go M_1^{\circ}=\go M_2^{\circ}$}}    
    \put(80.5,40){\contour{white}{$\go M_3^{\circ}$}}
	\put(18,72){\contour{white}{$\go m_1^{\circ}=\go m_2^{\circ}$}}
	\put(75,71){\contour{white}{$\go m_3^{\circ}$}}
	\end{small}
  \end{overpic}
}
\caption{(a) Sketch of a 3-legged spherical $3$-dof RPR manipulator. (b) Self-motion of type (I). (c) Self-motion of type (II). 
(d) 2-legged spherical manipulator (= spherical 4-bar mechanism).}
  \label{fig1}
\end{figure}

For the proof of Theorem \ref{thm4} (including the three additional cases $(\alpha,\beta,\gamma)$) we need the following result on 3-legged spherical 3-dof RPR manipulators 
proven in Lemma 2 of \cite{nawratil_proj} and Theorems 5 and 6 of \medskip \cite{nawratil_bond}:

\noindent
{\it
A 3-legged spherical 3-dof RPR manipulator with base anchor points $\go M_1^{\circ},\go M_2^{\circ},\go M_3^{\circ}$ and platform anchor points 
$\go m_1^{\circ},\go m_2^{\circ},\go m_3^{\circ}$ (see Fig.\ \ref{fig1}a) can only have a self-motion in the following two cases (under consideration of footnote 1): 
\begin{enumerate}[$(I)$]
\item
If $\go m_2^{\circ}=\go m_3^{\circ}$ coincides with $\go M_1^{\circ}$, then the platform can rotate freely about this point (see Fig.\ \ref{fig1}b). 
\item
If the base degenerates into one point; i.e. $\go M_1^{\circ}=\go M_2^{\circ}=\go M_3^{\circ}$. Then the manipulator can 
rotate in all poses about this point (see Fig.\ \ref{fig1}c).
\end{enumerate}
}

\subsubsection{$p=2$}
We can assume that $\go m_3,\go m_4,\go m_5$ are not collinear, 
as otherwise we get an architecturally singular pentapod. Therefore the platform can perform a pure spherical motion about 
the center $\go M_3=\go M_4=\go M_5$. Now the two legs $\go m_4\go M_4$ and $\go m_5\go M_5$ only imply one constraint 
in one of the following two cases:
\begin{enumerate}[1.]
\item
One anchor point is located in the center: It has to be a platform anchor point as otherwise we get an architecturally singular design. 
This implies item 1 of Theorem \ref{thm4}.
\item
The projection $\go m_4^{\circ}\go M_4^{\circ}$ and $\go m_5^{\circ}\go M_5^{\circ}$ of the legs $\go m_4\go M_4$ and $\go m_5\go M_5$ 
on the unit sphere centered in $\go M_3=\go M_4=\go M_5$ coincide. 
The condition $\go m_4^{\circ}=\go m_5^{\circ}$ implies that $\go M_3=\go M_4=\go M_5$ is located on the line spanned by $\go m_4$ and $\go m_5$. 
The condition $\go M_4^{\circ}=\go M_5^{\circ}$ implies the collinearity of all 5 base anchor points. 
Therefore we get the additional case $(\alpha)$.
\end{enumerate}
  
For the discussion of the case $p=3,4$ we can assume that no three base anchor points coincide, as otherwise we can only get a special case of $p=2$.

\subsubsection{$p=3$}
\begin{figure}[top] 
\centering
\subfigure[]{
 \begin{overpic}
    [width=35mm]{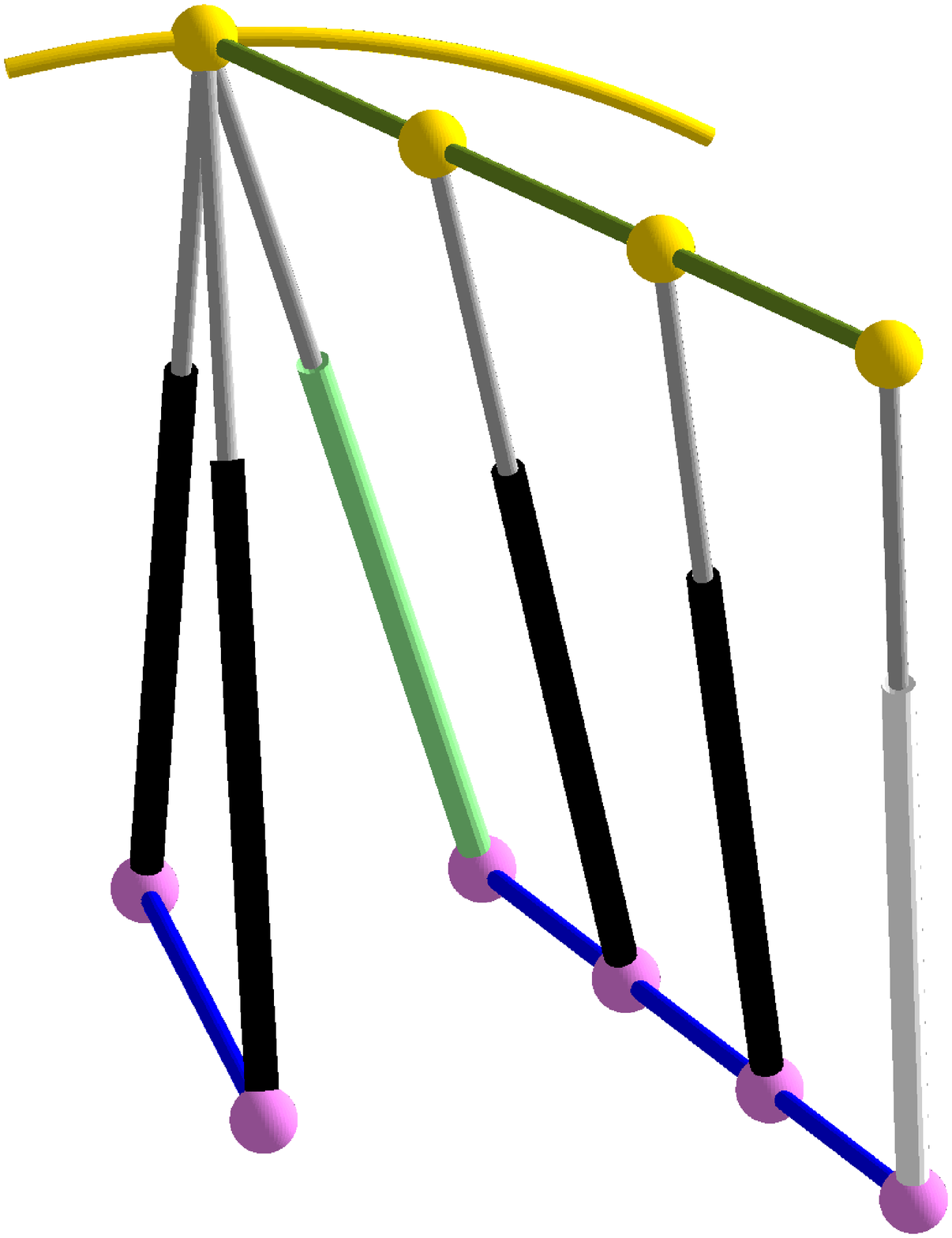}
  \begin{small} 
    \put(64,0){$\go M_1$}    
    \put(52,9){$\go M_2$}    
    \put(40,18){$\go M_3$}    
    \put(30.5,27){$\go M$}    
    \put(1,26){$\go M_5$}    
    \put(11,7){$\go M_4$}    
		\put(6,100){$\go m_4=\go m_5$}
		\put(25.5,84){$\go m_3$}
		\put(43.5,76){$\go m_2$}	
		\put(62.1,67){$\go m_1$}
		\put(54,93){$\go s$}	
	\end{small}
  \end{overpic}
  }
  \quad
\subfigure[]{
\begin{overpic}
    [width=34mm]{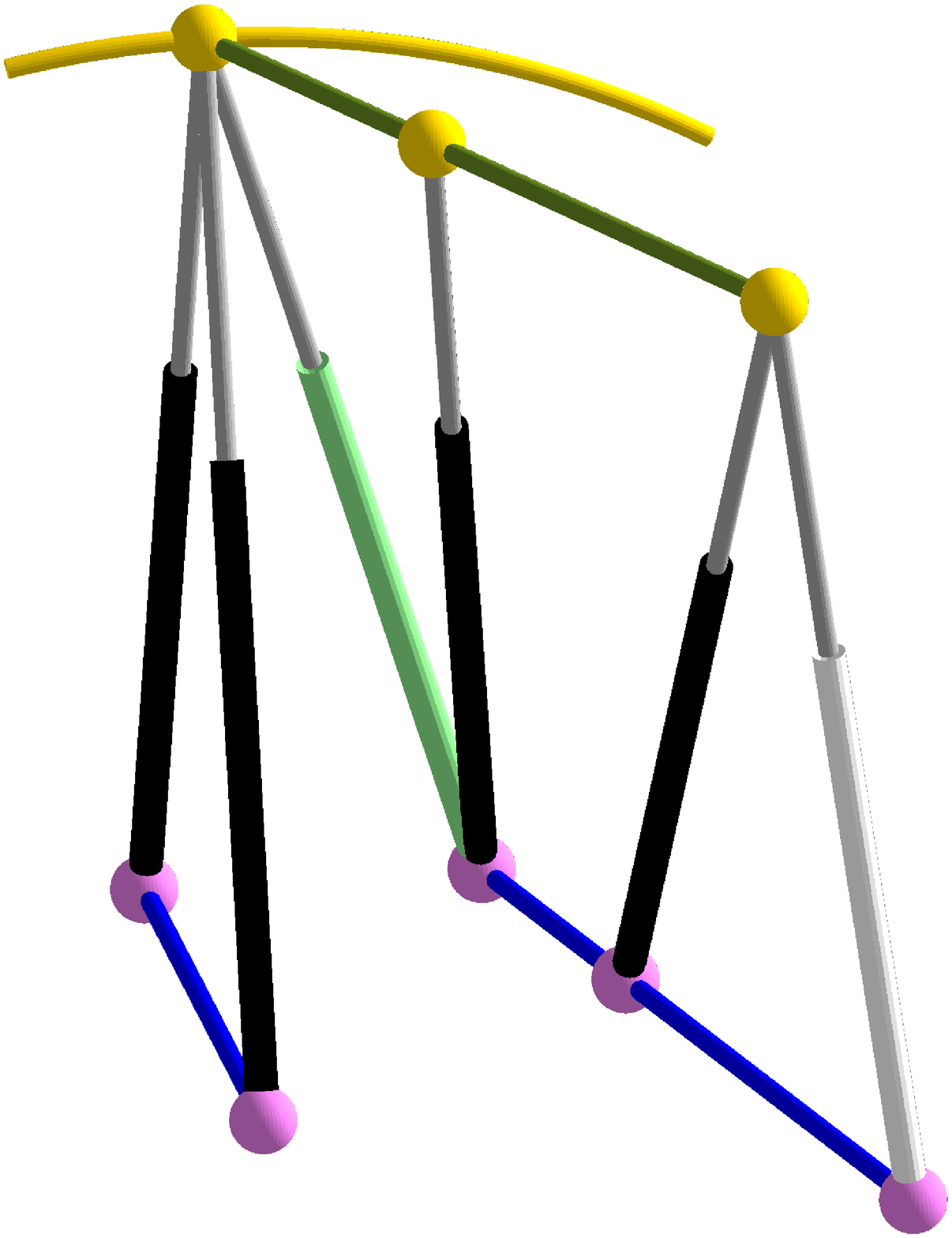}
	\begin{small}
    \put(63.5,0){$\go M_1$}    
    \put(40.5,17){$\go M_2$}    
    \put(29,26){$\go M_3$}    
    \put(1,26){$\go M_5$}    
    \put(11,7){$\go M_4$}    
		\put(6,100){$\go m_4=\go m_5$}
		\put(25.5,84){$\go m_3$}
		\put(56,80){$\go m_1=\go m_2$}
		\put(54,93){$\go s$}	
	\end{small}
  \end{overpic}
}
\quad
\subfigure[]{
\begin{overpic}
    [width=27mm]{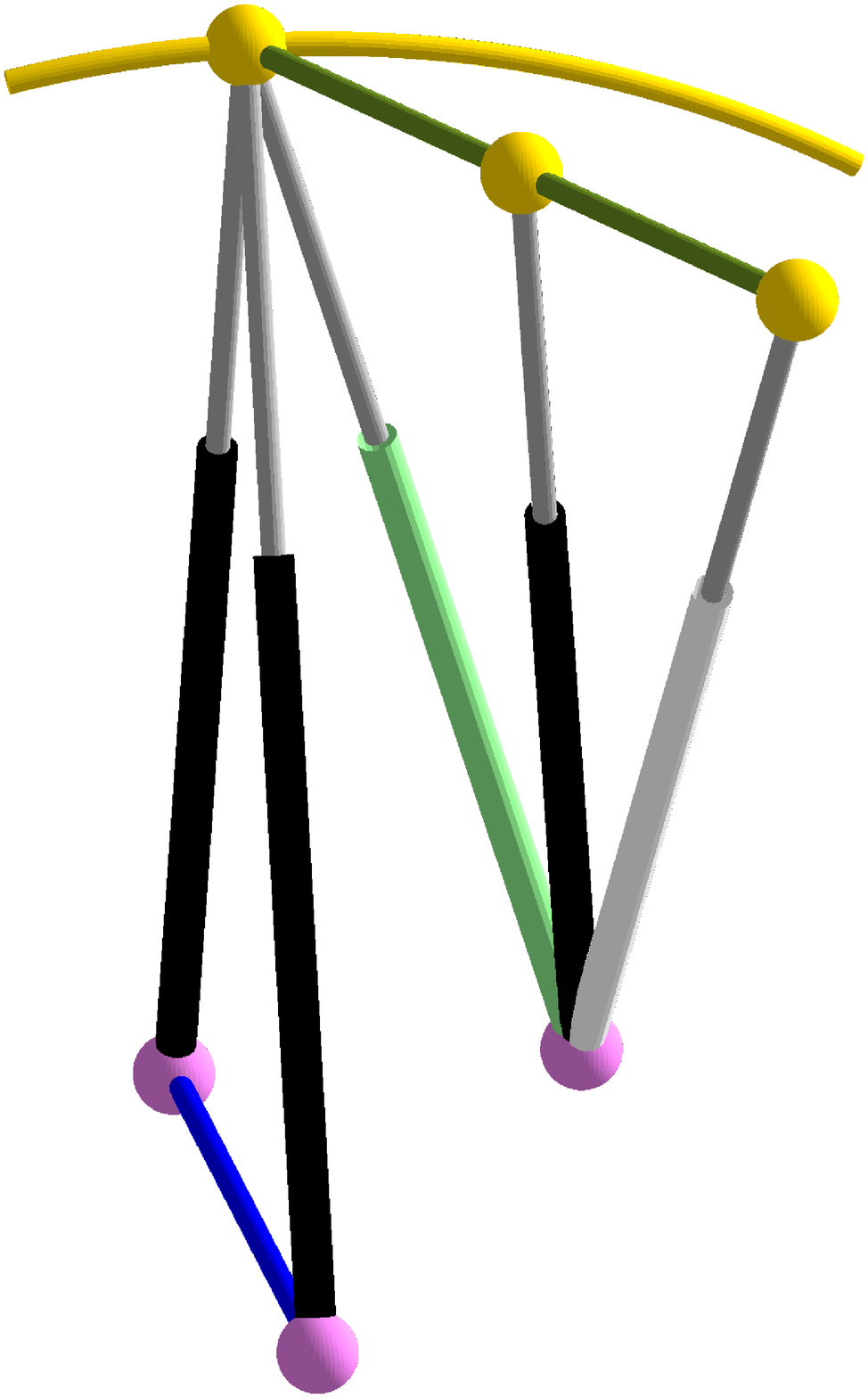}
	\begin{small}
    \put(30,17){$\go M_1=\go M_2$}    
    \put(1,23){$\go M_5$}    
    \put(11,4){$\go M_4$}    
		\put(6.5,100){$\go m_4=\go m_5$}
		\put(27,83.5){$\go m_2$}
		\put(60,71){$\go m_1$}
		\put(57,93){$\go s$}	
	\end{small}
  \end{overpic}
}
\quad
\subfigure[]{
\begin{overpic}
    [width=34mm]{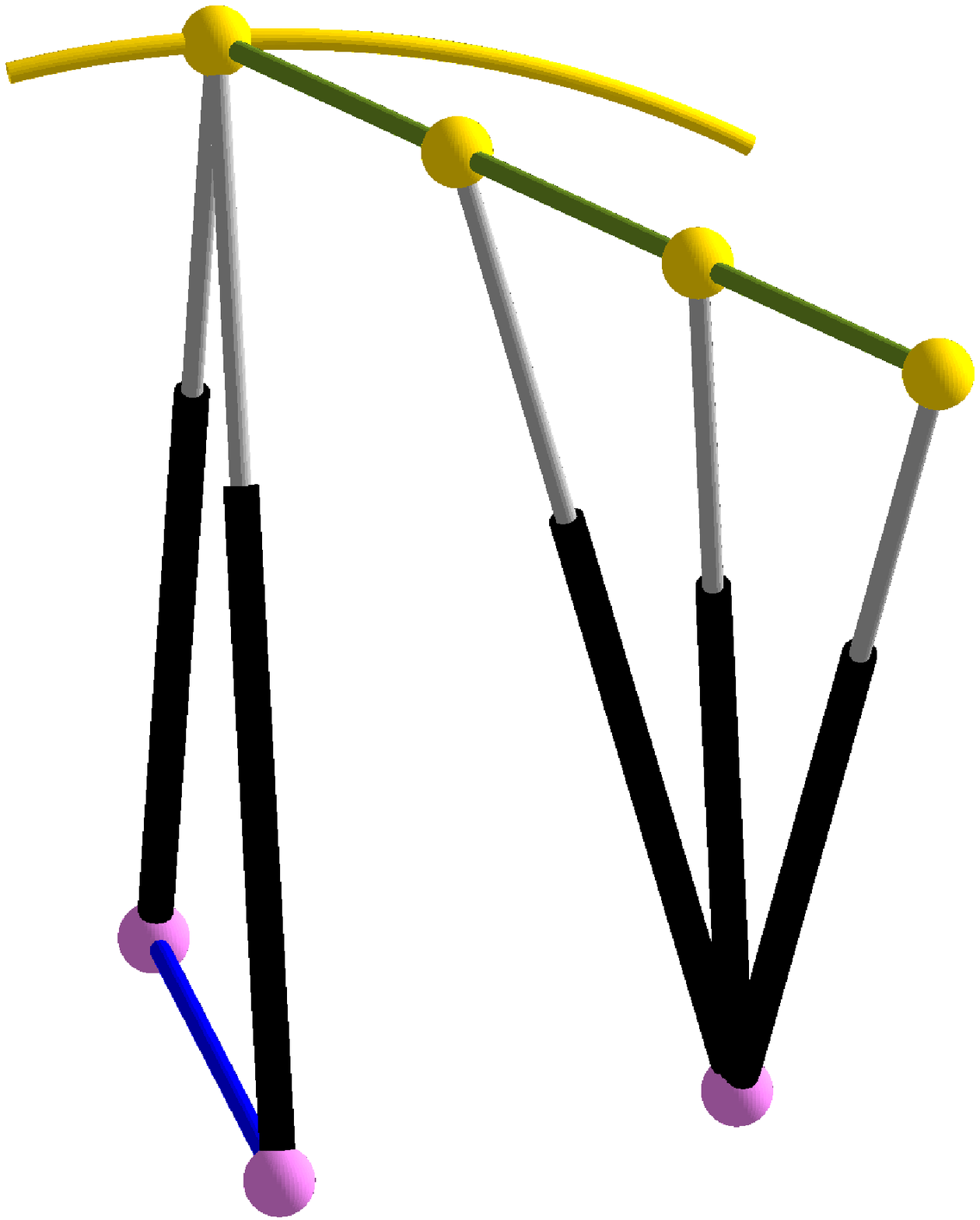}
	\begin{small}
    \put(40,1){$\go M_1=\go M_2=\go M_3$}    
    \put(1,23){$\go M_5$}    
    \put(11,4){$\go M_4$}    
		\put(6.5,100){$\go m_4=\go m_5$}
		\put(27,82.5){$\go m_3$}
		\put(46,75){$\go m_2$}
		\put(65,66.5){$\go m_1$}
		\put(57,93){$\go s$}	
	\end{small} 
  \end{overpic}
}
\caption{(a,b) Sketches of the two case listed in item 2(a)iiB of $p=3$. 
(c,d) Sketch of case 2(b) and 2(c), respectively, of $p=3$. 
}  
\label{fig2}
\end{figure}

For an intuitive argumentation of this case, we interchange platform and base; i.e.\ we have 
$\go M_1,\go M_2,\go M_3$ collinear, $\go m_4=\go m_5$ and the assumptions that neither 
$\go M_1,\ldots,\go M_5$ are collinear nor three platform anchor points coincide. 
As $\go m_4=\go m_5$ can only be located on a circle $\go s$ with axis $[\go M_4,\go M_5]$, we distinguish the 
following possibilities:
\begin{enumerate}[1.]
\item
$\go m_4=\go m_5$ is fixed during the 2-dimensional self-motion ($\Rightarrow$ spherical motion with center $\go m_4=\go m_5$): 
Now the three legs $\go m_i\go M_i$ ($i=1,2,3$) are only allowed to imply one constraint. We have to discuss the following three possibilities: 
	\begin{enumerate}[(a)]
	\item  
	If $\go M_1=\go M_2$ is located in the center $\go m_4=\go m_5$. This implies item 2(a) of Theorem \ref{thm4}.
	\item
	If $\go M_1$ is located in the center and  the projection $\go m_2^{\circ}\go M_2^{\circ}$ and $\go m_3^{\circ}\go M_3^{\circ}$ 
	of the remaining two legs on the unit sphere centered in $\go m_4=\go m_5$ coincide; i.e.\ $\go m_2^{\circ}=\go m_3^{\circ}$ and $\go M_2^{\circ}=\go M_3^{\circ}$.
	This implies $\go m_2,\go m_3, \go m_4=\go m_5$ collinear and we get item 2(b) of Theorem \ref{thm4}.
	\item
	The projection $\go m_i^{\circ}\go M_i^{\circ}$ of the three legs on the unit sphere centered in $\go m_4=\go m_5$ coincide. 
	For the same reason as in the last item $\go m_1,\go m_2,\go m_3, \go m_4=\go m_5$ have to be collinear 
	as well as $\go M_1,\go M_2,\go M_3$. This yields the additional case $(\beta)$.
	\end{enumerate}
\item
$\go m_4=\go m_5$ moves on $\go s$: In each pose of $\go m_4=\go m_5$ there has to be 
a 1-dimensional spherical self-motion with center $\go m_4=\go m_5$. 
Now it is not possible that one of the $\go M_i$'s ($i=1,2,3$) always coincide with $\go m_4=\go m_5$ 
as this point moves along $\go s$. Therefore we can consider the resulting spherical manipulator 
$\go m_i^{\circ}\go M_i^{\circ}$ for $i=1,2,3$ with respect to the unit sphere centered in $\go m_4=\go m_5$. 
The resulting spherical manipulator is:
\begin{enumerate}[(a)]
\item
3-legged: Due to the cited results (at the beginning of Section \ref{sec:pthm4}) a necessary condition for its self-mobility is that 
either two platform points or two base points coincide:
	\begin{enumerate}[i.]
	\item
	$\go M_1^{\circ}=\go M_2^{\circ}$: This is only possible for all poses of $\go m_4=\go m_5$ on $\go s$ 
	if $\go M_1=\go M_2$ holds. Now we have to distinguish between the self-motions (I) and (II): 
		\begin{enumerate}[A.]
		\item
		$\go M_1^{\circ}=\go M_2^{\circ}$ coincides with $\go m_3^{\circ}$:
		We get the 1-dimensional self-motion if $\go m_3,\go m_4=\go m_5$ and $\go M_1=\go M_2$ are collinear. 
		Due to the fixed leg length the distance from $\go M_1=\go M_2$ to $\go m_4=\go m_5$ has to be constant during the 
		motion of $\go m_4=\go m_5$ on $\go s$. Therefore $\go M_1=\go M_2$ has to be located on the axis $[\go M_4,\go M_5]$. 
		Moreover as the distance from $\go m_3$ to $\go m_4=\go m_5$ is trivially fixed also $\go M_3$ has to be located on $[\go M_4,\go M_5]$. 
		Therefore all base anchor points have to be collinear; a contradiction.
		\item
		$\go M_1^{\circ}=\go M_2^{\circ}=\go M_3^{\circ}$: Therefore $\go M_1=\go M_2=\go M_3$ has to hold. 
		As a consequence $\go s$ has to be a circle on a sphere centered in  $\go M_1=\go M_2=\go M_3$, which implies 
		the collinearity of $\go M_1=\go M_2=\go M_3,\go M_4,\go M_5$; a contradiction. 
		\end{enumerate}
	\item
	$\go m_1^{\circ}=\go m_2^{\circ}$: This implies the collinearity of $\go m_1,\go m_2, \go m_4=\go m_5$. 
	Now we have to distinguish between the self-motions (I) and (II):
		\begin{enumerate}[A.]
		\item
		$\go m_1^{\circ}=\go m_2^{\circ}$ coincides with $\go M_3^{\circ}$:
		We get the 1-dimensional self-motion if $\go m_1,\go m_2, \go m_4=\go m_5$ and $\go M_3$ are collinear. 
		Due to the fixed leg length the distance from $\go M_3$ to $\go m_4=\go m_5$ has to be constant during the 
		motion of $\go m_4=\go m_5$ on $\go s$. Therefore $\go M_3$ has to be located on the axis $[\go M_4,\go M_5]$. 
		Moreover as the distance from $\go m_i$ ($i=1,2$) to $\go m_4=\go m_5$ is trivially fixed also $\go M_i$ has to be located on $[\go M_4,\go M_5]$. 
		Therefore all base anchor points have to be collinear; a contradiction.
		\item
		$\go m_1^{\circ}=\go m_2^{\circ}=\go m_3^{\circ}$: Therefore $\go m_1,\go m_2, \go m_3,\go m_4=\go m_5$ are collinear. 
		As no three platform anchor points are allowed to coincide we remain with two possibilities:
			\begin{itemize}
			\item[$\bullet$]
			$\go m_1,\go m_2, \go m_3$ are pairwise distinct: Therefore the first three legs of the pentapod span a regulus $\mathcal{R}$ of lines. 
			It is well known (e.g.\ \cite{borras_ark}) that each line of $\mathcal{R}$ can be replaced by any other line of $\mathcal{R}$ 
			without changing the singularity-set of the pentapod. E.g.\ we can replace the first leg by the line of $\mathcal{R}$ through $\go m_4=\go m_5$ (see Fig.\ \ref{fig2}a). 
			Its base anchor point  $\go M$ cannot be located on the axis $[\go M_4,\go M_5]$, as otherwise the originally given pentapod is an architecturally singular one 
			(special case of item 10 of Theorem 3 given in \cite{kargernonplanar}; see also item 9 of Corollary 1 given in \cite{ns}). 
			Therefore this already shows that $\go m_4=\go m_5$ is fixed, which contradicts our assumption that the point moves along $\go s$. 
			\item[$\bullet$]
			$\go m_1,\go m_2, \go m_3$ are not pairwise distinct: W.l.o.g.\ we can assume $\go m_1=\go m_2$. Now
			$\go M_3$ cannot be located on the axis $[\go M_4,\go M_5]$, as otherwise the pentapod is architecturally singular 
			(item 7 of Theorem 3 of \cite{kargernonplanar}). Now we can perform the following singular-invariant leg-rearrangement according to \cite{borras_ark}: 
			We can replace the leg $\go m_1\go M_1$ by the leg $\go m_4\go M_3$ (see Fig.\ \ref{fig2}b). This again shows that $\go m_4=\go m_5$ is fixed; a contradiction.
			\end{itemize}
		\end{enumerate}
	\end{enumerate}
\item
2-legged: W.l.o.g.\ we can assume $\go M_1^{\circ}=\go M_2^{\circ}$ and $\go m_1^{\circ}=\go m_2^{\circ}$ (see Fig.\ \ref{fig1}d). 
This implies $\go M_1=\go M_2$ and the collinearity of $\go m_1,\go m_2, \go m_4=\go m_5$. 
$\go M_1=\go M_2$ cannot be located on the axis $[\go M_4,\go M_5]$, as otherwise we get an architecturally singular pentapod 
(special case of item 8 of Theorem 3 given in \cite{kargernonplanar}). Now we can perform the following singular-invariant 
leg-rearrangement according to \cite{borras_ark}: We can replace the leg $\go m_1\go M_1$ by the leg $\go m_4\go M_1$ (see Fig.\ \ref{fig2}c). 
This already shows that $\go m_4=\go m_5$ is fixed; a contradiction. 
\item
1-legged: From $\go m_1^{\circ}=\go m_2^{\circ}=\go m_3^{\circ}$ we get the collinearity of $\go m_1,\go m_2, \go m_3,\go m_4=\go m_5$. 
Moreover $\go M_1^{\circ}=\go M_2^{\circ}=\go M_3^{\circ}$ implies $\go M_1=\go M_2=\go M_3$ and therefore the first three legs are 
in a pencil of lines (see Fig.\ \ref{fig2}d). This is an architecturally singular case (item 6 of Theorem 3 given in \cite{kargernonplanar}).
\end{enumerate}
\end{enumerate}

\subsubsection{$p=4$}

For the discussion of the case $\go m_1,\ldots ,\go m_4$ collinear we can assume that $\go M_5$ does not coincide 
with another base anchor point, as otherwise we can only get a special case of $p=3$. 
If we remove the fifth leg, then we get a quadropod, which can have a 2-dimensional self-motion or a higher one: 
	\begin{enumerate}[1.]
	\item
	2-dimensional self-motion: Now this motion consists of a 1-dimensional rotation of the platform about the 
	carrier line $\go g$ of $\go m_1,\ldots ,\go m_4$ and a 1-dimensional motion of $\go g$ itself. 
	As we can assume that $\go m_5$ is not located on $\go g$, the point $\go M_5$ has to be located on $\go g$ in each pose of the self-motion. 
	Moreover as the leg lengths is fixed $\go M_5$  has to coincide with the same point $\go G\in\go g$. 
	  
	Now it remains to guarantee the 1-dimensional motion of $\go g$, which can only be a spherical one due to $\go M_5=\go G$ (center of the spherical motion). 
		\begin{enumerate}[(a)]
		\item
		If $\go G\neq\go m_i$ holds for all $i\in\left\{1,2,3,4\right\}$, then there can only be a 
		motion of $\go g$ if $\go M_1,\ldots ,\go M_5$ are collinear. This yields the additional case $(\gamma)$.
		\item
		If $\go G=\go m_1$ holds, then only  $\go M_2,\ldots ,\go M_5$ have to be collinear. This yields item 3 of Theorem \ref{thm4}.
		\item
		If $\go G=\go m_1=\go m_2$ holds, then only  $\go M_3,\ldots ,\go M_5$ have to be collinear. This yields again item 2(b) of Theorem \ref{thm4}.
		\item
		If $\go G=\go m_1=\go m_2=\go m_3$ holds. This yields again item 1 of Theorem \ref{thm4}.
		\end{enumerate}
	\item
	higher-dimensional self-motion: The quadropod can only have an $n$-dimensional self-motion with $n>2$ in one of the 
	following cases (according to \cite{nawratilmulti}) if it is not architecturally singular:
		\begin{enumerate}[(a)]
		\item
		If $\go m_1=\go m_2=\go m_3$ coincides with $\go M_4$: This case implies once more item 1 of Theorem \ref{thm4}.
		\item
		If $\go m_1=\go m_2$ coincides with $\go M_3=\go M_4$: This case implies again item 2(a) of Theorem \ref{thm4} (with the extra condition 
		$\go m_1,\ldots ,\go m_4$ collinear). This finishes the discussion of item (c) of Theorem \ref{thm1a}. 
				\end{enumerate}
	\end{enumerate}


\subsection{Study of item (d) of Theorem \ref{thm1a}}\label{part(d)}

Due to the discussion done in Section \ref{part(d)}, we can assume the following for the pentapod design 
given in item (d) of Theorem \ref{thm1a}:
$\go M_4\neq\go M_5$, $\go m_4\neq\go m_5$, and neither $\go M_1=\go M_2=\go M_3$ nor $\go m_1=\go m_2=\go m_3$ holds. 

Moreover the first, second and third leg span a regulus  $\mathcal{R}$ (which can degenerate into two pencils of lines). 
W.l.o.g.\ we can assume that $\go M_1,\go M_2,\go M_3$ are pairwise distinct, as otherwise we can replace one of the 
three legs by a line of  $\mathcal{R}$ without changing the singularity-set of the pentapod (cf.\ \cite{borras_ark}). 
Due to this possibility of leg-replacements within $\mathcal{R}$ we have even a free choice of pairwise distinct anchor points $\go M_1,\go M_2,\go M_3$ on the line $\go g$. 

Moreover the distance between $\go g$ and $\go h$ can be assumed to be $1$, which eliminates the scaling factor.
Therefore the base anchor points of the pentapod can be coordinatized as follows:
\begin{align*}
\Vkt M_1&=(0,0,0)^T, &\quad \Vkt M_2&=(0,1,0)^T, &\quad \Vkt M_3&=(0,-1,0)^T, \\
\Vkt M_4&=(1,0,0)^T, &\quad \Vkt M_5&=(1,B,0)^T. &\quad &\phm
\end{align*}
The platform has the following coordinatization: 
\begin{align*}
\Vkt m_1&=(0,0,0)^T, &\quad \Vkt m_2&=(0,b_2,0)^T, &\quad \Vkt m_3&=(0,b_3,0)^T, \\
\Vkt m_4&=(a,b_4,0)^T, &\quad \Vkt m_5&=(a,b_5,0)^T, &\quad &\phm
\end{align*}
where $a>0$ can be assumed w.l.o.g.. Note that $aB(b_4-b_5)\neq 0$ holds and that $b_2=b_3=0$ yields a contradiction. 

The 2-dimensional self-motion $\mathcal{S}$ of this pentapod can be assumed to be of type $\beta=1$, as 
the existence of a translational self-motions implies an affine coupling of the 
planar platform and the planar base (see Fig.\ \ref{figextra} under consideration of the last paragraph of Section \ref{sec:def}), 
which was already discussed in Section \ref{sec:pthm3}. 

The triangles $\go M_1,\go M_4,\go M_5$ and $\go m_1,\go m_4,\go m_5$ determine uniquely   
a regular affinity $\mu$: $\go M_i\mapsto\go m_i$ for $i=1,4,5$. 
As not $\mu(\go M_2)=\go m_2$ and $\mu(\go M_3)=\go m_3$ can hold simultaneously (otherwise we 
get again an planar affine pentapod) we can assume w.l.o.g.\ that $\mu(\go M_2)\neq \go m_2$ holds, which 
equals the condition:
\begin{equation}\label{ass:nonaff}
Bb_2+b_4-b_5\neq 0.
\end{equation}
Based on this preparatory work/assumptions we prove by direct computations that 
the set of projected bonds $\mathcal{B}_f$ cannot be 1-dimensional, which shows that item (d) of Theorem \ref{thm1a} 
contains no further solutions beside the affine ones listed in Theorem \ref{thm3}.
This more technical part of the proof is given in the Appendix in order to streamline the presentation and 
to improve the readability of the paper. \hfill $\BewEnde$


\section{Conclusions for hexapods}\label{sec:hexapod}

The adding of an arbitrary leg to any pentapod with mobility 2 yields in the 
generic case a hexapod with mobility 1. Now the question arises for those cases where the attachment does not restrict the dimension of the mobility. 
Moreover we are only interested in non-architecturally singular hexapods as they have a practical application in robotics as so-called  
Stewart-Gough platforms. In the following we give a complete list of these manipulators:

\begin{thm}\label{thm6}
A non-architecturally singular hexapod with mobility 2 belongs to one of the following six cases (under consideration of footnote 1): 
\begin{enumerate}
\item
Platform and base are congruent: We get the trivial 2-dimensional translation of the congruent hexapod. 
\item
$\go m_1,\ldots ,\go m_5$ collinear and $\go M_2,\ldots ,\go M_6$ collinear: 
We get a 2-dimensional spherical self-motion if $\go m_1$ coincides with $\go M_6$.
\item
$\go m_1=\go m_2, \go m_3,\go m_4 ,\go m_5$ collinear and $\go M_3,\ldots ,\go M_6$ collinear: 
We get a 2-dimensional spherical self-motion if $\go m_1=\go m_2$ coincides with $\go M_6$.
\item
$\go m_1=\go m_2=\go m_3,\go m_4 ,\go m_5$ collinear and $\go M_4,\go M_5 ,\go M_6$ collinear: 
We get a 2-dimensional spherical self-motion if $\go m_1=\go m_2=\go m_3$ coincides with $\go M_6$.
\item
$\go m_1=\go m_2, \go m_3,\go m_4$ collinear and $\go M_3,\go M_4,\go M_5=\go M_6$ collinear: 
We get a 2-dimensional spherical self-motion if $\go m_1=\go m_2$ coincides with $\go M_5=\go M_6$.
\item
$\go m_1=\go m_2=\go m_3$ and $\go M_5=\go M_6$: 
We get a 2-dimensional spherical self-motion if $\go m_1=\go m_2=\go m_3$ coincides with $\go M_5=\go M_6$.
\end{enumerate}
\end{thm}

\noindent
{\sc Proof:}
Clearly architecturally singular pentapods yield architecturally singular hexapods by the attachment of an arbitrary leg. 
Therefore we can focus on non-architecturally singular pentapods with mobility 2. 
They are given in the Theorems \ref{thm2}, \ref{thm3} and \ref{thm4} or belong to the case $p=5$ of item (c) of Theorem \ref{thm1a}. 

As the 2-dimensional translation of the congruent hexapod is trivial we can proceed with item 2 of Theorem \ref{thm3}. 
We assume that our hexapod possesses the 2-dimensional Sch\"onflies self-motion described in item 2 of Theorem \ref{thm3}. 
If we take any of the six legs away we have to end up with the pentapod of item 2 of Theorem \ref{thm3} up to permutation of indices. 
As a consequence the six anchor points have to be located on two parallel lines, where each line carrier three anchor points. 
Therefore the anchor points are located on a degenerated conic, which already shows that the hexapod is architecturally singular. 

In the following we investigate the cases listed in Theorem \ref{thm4}. We start with item 1; i.e.\ $\go M_3=\go M_4=\go M_5$:  
We consider the 2-dimensional self-motion of this pentapod when $\go M_3=\go M_4=\go M_5$ coincides with $\go m_1$. 
This spherical self-motion is not restricted by an additional leg $\go m_6\go M_6$ in one of the following cases: 
	\begin{enumerate}[$\star$]
	\item
	$\go M_6=\go M_3=\go M_4=\go M_5$: This yields an architecturally singular hexapod.
	\item 
	$\go m_6=\go m_1$: This yields item 6 of Theorem \ref{thm6}.
	\item
	$\go M_3=\go M_4=\go M_5,\go M_2,\go M_6$ collinear and $\go m_1,\go m_2,\go m_6$ collinear: 
	This yields item 4 of Theorem \ref{thm6}. 
	\end{enumerate}
The discussion for the items 2(a), 2(b) and 3 of Theorem \ref{thm4} can be done analogously. 
They already imply all the remaining cases given in Theorem \ref{thm6}. 

Therefore we are left with the pentapods belonging to the case $p=5$ of item (c) of Theorem \ref{thm1a}.
Due to Corollary \ref{thm1b} and the list of architecturally singular 
pentapods (cf.\ Theorem 3 of \cite{kargernonplanar} under consideration of \cite{nawratilpenta}) 
there does not exist a pentapod of this type with mobility 3 (or higher), 
which is not architecturally singular. Therefore all non-architecturally singular pentapods of this type have mobility 2, where 
one degree of freedom is the rotation about the carrier line $\go g$ of $\go m_1,\ldots ,\go m_5$. 
This mobility is not restricted by the attachment of a sixth leg only in the already obtained 
five cases 2--6 of Theorem \ref{thm6}. 
This can easily be seen by performing analogous considerations as in the proof of Theorem \ref{thm4}, case $p=4$, item 1. \hfill $\BewEnde$

\begin{rmk}
Note that the items 1 and 6  of Theorem \ref{thm6} were already listed in \cite{nawratilmulti}. 
The remaining four cases are new to the best knowledge of the authors. 
Moreover this listing also shows that there does not exist a non-architecturally singular hexapod 
with a 2-dimensional self-motion of type $\beta=0$, which was also an open question of \cite{nawratilmulti}. \hfill $\diamond$
\end{rmk}

Due to Theorem \ref{thm6} and the results of \cite{nawratilmulti} only the classification of all non-architecturally 
singular hexapods with a 1-dimensional self-motion remains for the complete solution of the famous Borel-Bricard problem. 
Some necessary conditions for these overconstrained mechanisms were already presented in \cite{gns1}. 
Further investigations on this topic are dedicated to future research.


\section*{Acknowledgments}
The first author's research is funded by the Austrian Science Fund (FWF): P24927-N25 - ``Stewart Gough platforms with self-motions''. 
The second author's research is supported by the Austrian Science Fund (FWF): W1214-N15/DK9 and P26607 - ``Algebraic Methods in Kinematics: Motion Factorisation and Bond Theory''.

\section*{Appendix} 

This technical part of the proof started in Section \ref{part(d)} splits up into a general case and a special one. 
We recommend to study them together with the corresponding Maple worksheets, which 
can be downloaded from the homepage ({\tt http://www.geometrie.tuwien.ac.at/nawratil}) of the first author.

\subsection*{General case: $e_0e_1(a-1)-e_2e_3(a+1)\neq 0$}

Due to this assumption and Eq.\ (\ref{ass:nonaff}) we can solve 
$\Psi,\Delta_{2,1},\Delta_{4,1},\Delta_{5,1}$ for $f_0,\ldots ,f_3$ w.l.o.g.. Then we plug the obtained 
solutions into the remaining two equations $\Lambda_1$ and $\Delta_{3,1}$. The numerators of the 
resulting expressions are denoted by $G_1[66139]$ and $G_3[160]$, where the number in the brackets gives the number of 
terms. Note that $G_1=0$ represents a surfaces of degree 8 in the Euler parameter space $P^3$, in contrast to $G_3=0$ and $N=0$ which 
are quadrics. 

Then we eliminate $e_0$ by computing the resultant $H_j$ of $G_j$ and $N$ with respect to $e_0$ for $j=1,3$. 
We get $H_3[170]$ and $H_1$ is a perfect square; i.e.\ $H_1=K[380]^2$. Moreover in $H_3$ and $K$ only 
even powers of $e_3$ appear and therefore we can substitute $e_3^2$ by $\overline{e}_3$ which yields $\overline{H}_3$ and $\overline{K}$, respectively. 
Finally we compute the resultant of $\overline{H}_3$ (quadratic in $\overline{e}_3$) and $\overline{K}$ (quartic in $\overline{e}_3$) 
with respect to  $\overline{e}_3$ which yields $L[167161]$. 

A necessary condition for the existence of a common curve of the octic surface and the two quadrics ($\Leftrightarrow$ $\mathcal{B}_f$ is 1-dimensional $\Leftrightarrow$ $\beta=1$) 
is that $L$ is fulfilled identically for all $e_1,e_2$. As $L$ factors into 
\begin{equation*}
L=e_1^4e_2^4a^2B(b_4-b_5)(Bb_2+b_4-b_5)^2U[497]V[5030],
\end{equation*}
where $V=0$ is pseudo-solution implied by the elimination process\footnote{This can easily be checked by 
computing the resultant $G_{1,3}$ of $G_1$ and $G_3$ with respect to $e_0$, which is of degree 8 in $\overline{e}_3$. 
Then the resultant of $G_{1,3}$ and $H_3$ (or $K$) with respect to $\overline{e}_3$ does not contain the factor $V$.}, 
we remain with $U=0$, which is a homogeneous quartic equation in $e_1,e_2$. We denote the coefficient of $e_1^{i}e_2^{j}$ of $U$ by $U_{i,j}$ and 
end up with the following three conditions: 
\begin{equation*}
U_{4,0}=U_{04}=(b_2+b_3)^2W_1[44]=0, \qquad U_{3,1}=-U_{1,3}=(b_2+b_3)W_2[81]=0,\qquad U_{2,2}=W_3[125]=0.
\end{equation*}
For $b_2=-b_3$ the condition $W_3=0$ simplifies to $b_2^2(Bb_2+b_4-b_5)^2=0$, which cannot vanish w.c.. 
Therefore we can assume $b_2\neq -b_3$, which implies $W_1=W_2=W_3=0$.
Now we eliminate $b_5$ by calculating the resultant $E_k$ of $W_i$ and $W_j$ with respect to $b_5$ for pairwise distinct 
$i,j,k\in\left\{1,2,3\right\}$. Then the greatest common divisor $GCD$ of $E_1,E_2,E_3$ equals 
$aB^4(b_2+b_3)^2b_2b_3(b_2-b_3)T$ with
\begin{equation*}
T=(a^2+b_4^2)(b_2^2+b_3^2)-2ab_2b_3(b_2-b_3)-2b_2b_3b_4(b_2+b_3)+2b_2^2b_3^2.
\end{equation*}
In the following case study we show that the factors of $GCD$ does not imply a solution to our problem: 
\begin{enumerate}[1.] 
\item
$b_2=0$: Now $W_1$ can only vanish w.c.\ for: 
	 \begin{enumerate}[(a)]
	 \item
	 $b_5=-a$: Then $W_2=0$ implies $b_4=-Ba-a$. Finally $W_3$ equals $B^2a^3b_3^4(B^2+2B+2)$ 
	 which cannot vanish w.c.\ over $\RR$. 
	 \item
	 $b_4=-Ba-a$ and $b_5\neq -a$: Now $W_2=0$ already yields the contradiction.
	 \end{enumerate}
\item
Analogously the cases $b_3=0$ and $b_2=b_3$ can be studied, as they are geometrically identical to the case $b_2=0$ 
($\mathcal{R}$ splits up into two pencils of lines). As they also yield no solution, we can assume 
$b_2b_3(b_2-b_3)\neq 0$ for the discussion of the remaining case $T=0$.
\item
$T=0$: We can solve $T=0$ w.l.o.g.\ for $a$ which yields:
\begin{equation*}
a=\frac{b_2b_3(b_2-b_3)\pm i\left[
b_2^2(b_3-b_4)+b_3^2(b_2-b_4)\right]}
{b_2^2+b_3^2}.
\end{equation*}  
As $a$ has to be a real number the expression in the brackets has to vanish, which implies 
$b_4=\frac{b_2b_3(b_2+b_3)}{b_2^2+b_3^2}$.
Then the common factors of $W_1,W_2,W_3$ are $b_2b_3(b_2-b_3)J[45]$. $J$ is quadratic in $b_5$ and can be solved for 
this unknown w.l.o.g.. Now $b_5$ can only be real if $B=2\frac{b_2-b_3}{b_2+b_3}$ holds which yields
\begin{equation*}
b_5=\frac
{(3b_2^2-2b_2b_3+3b_3^2)b_2b_3}
{(b_2+b_3)(b_2^2+b_3^2)}.
\end{equation*}
In this case $\overline{H}_3$ only depends on $e_1$ and $e_2$ (and not longer on $\overline{e}_3$) and reads as 
$e_1^2e_2^2b_2^2b_3^2(b_2+b_3)^{-2}(b_2-b_3)^4$. Therefore $e_1=0$ or $e_2=0$ has to hold. 
In both cases $\overline{K}=0$ implies $b_2^2-b_2b_3+2b_3^2=0$, which has no real solution.
\end{enumerate}
Due to this discussion we can assume $GCD\neq 0$ and therefore we can delete all factors of $E_1,E_2,E_3$, which are also contained within $GCD$. 
We denote the remaining expressions by ${E}_1^*[464]$, ${E}_2^*[271]$ and ${E}_3^*[95]$, where ${E}_3^*$ can be factored but this 
is not of importance for that what follows.  We compute the resultant $D_k$ of $E_i^*$ and $E_j^*$ with respect to $B$ for pairwise distinct 
$i,j,k\in\left\{1,2,3\right\}$. As all common factors of $D_1,D_2,D_3$ imply $GCD=0$, we get a contradiction.

\subsection*{Special case: $e_0e_1(a-1)-e_2e_3(a+1)= 0$}

\begin{enumerate}[1.]
\item
$e_2\neq 0$: Under this assumption we can solve $e_0e_1(a-1)-e_2e_3(a+1)= 0$ for $e_3$ w.l.o.g.. 
		\begin{enumerate}[(a)]
		\item
		$\mu$ is no orientation preserving congruence transformation: Under this assumption we an solve 
		$\Psi,\Delta_{4,1},\Delta_{5,1}$ for $f_1,f_2,f_3$ w.l.o.g.. Then we plug the obtained 
		solutions into $\Delta_{2,1}$ and $\Delta_{3,1}$. The numerators of the 
		resulting expressions are denoted by $G_2[865]$ and $G_3[865]$, which are both of degree 6 in $e_0,e_1,e_2$. 
		Then we eliminate $e_0$ by computing the resultant $H_j$ of $G_j$ and $N$ with respect to $e_0$ for $j=2,3$. 
		These expressions factor into $H_i=a^2e_1^2e_2^2K_i[54]^2$. Therefore both expressions $K_2$ and $K_3$ have to be fulfilled 
		independently of $e_1,e_2$. 
		It can easily be seen that the resulting system of equations has no solution without yielding a contradiction. 
		\item
		$\mu$ is an orientation preserving congruence transformation; i.e.\ $a=1$, $b_4=0$ and $b_5=B$: 
		Then $\Psi,\Delta_{2,1},\Delta_{5,1}$ can be solved for $f_1,f_2,f_3$ w.l.o.g.. We plug the obtained 
		solutions into $\Delta_{4,1}$ and denote the numerator by $G_4[14]$, which is again of degree 6 in $e_0,e_1,e_2$. 
		Then we eliminate $e_0$ by computing the resultant $H_4$ of $G_4$ and $N$ with respect to $e_0$ which yield
		$B^2e_1^2e_2^2(Be_1-e_2)^2$. This expression cannot vanish w.c.. 
	 	\end{enumerate}	
\item
$e_2= 0$: This implies $a=1$. We distinguish the following two subcases:
		\begin{enumerate}[(a)]
		\item
		$\mu$ is no orientation reversing congruence transformation: The discussion can be done similarly to item 1(a).
		\item
		$\mu$ is an orientation reversing congruence transformation; i.e.\ $b_4=0$ and $b_5=-B$: The discussion can be done similarly to 
		item 1(b). This closes the proof of Theorem \ref{thm4}. \hfill $\BewEnde$
	 	\end{enumerate}	
\end{enumerate}

\end{document}